\documentclass[runningheads]{llncs}

\usepackage{eccv}

\usepackage{eccvabbrv}

\usepackage{graphicx}
\usepackage{booktabs}

\usepackage[accsupp]{axessibility}  

\usepackage[pagebackref,breaklinks,colorlinks,citecolor=eccvblue]{hyperref}

\usepackage{orcidlink}

\usepackage{enumitem}
\usepackage{colortbl}

\begin{document}

\title{VisionGPT: Vision-Language Understanding Agent Using Generalized Multimodal Framework} 

\titlerunning{VisionGPT}

\author{%
\begin{tabular}{ccc}
Chris Kelly\thanks{Authors contributed equally.} & Luhui Hu\textsuperscript{$\star$} & Bang Yang\textsuperscript{$\star$} \\
\textit{Stanford University} & \textit{Seeking AI} & \textit{Peking University} \\
\email{ckelly24@stanford.edu} & & \email{yangbang@pku.edu.cn} \\[2ex]
Yu Tian & Deshun Yang & Cindy Yang \\
\textit{Harvard University} & \textit{Seeking AI} & \textit{University of Washington, Seattle} \\
\email{ytian11@meei.harvard.edu} & & \email{c1ndyy@uw.edu} \\[2ex]
Zaoshan Huang & Zihao Li & Jiayin Hu \\
\textit{Seeking AI} & \textit{Seeking AI} & \textit{University of California, Los Angeles} \\
 & \email{li981354@seeking.ai} & \email{greenlake@ucla.edu} \\[2ex]
 & Yuexian Zou & \\
 & \textit{Peking University} & \\
 & \email{zouyx@pku.edu.cn} &
\end{tabular}
}

\institute{} 

\authorrunning{C. Kelly, L. Hu, B. Yang et al.}




\maketitle


\begin{abstract}
With the emergence of large language models (LLMs) and vision foundation models, how to combine the intelligence and capacity of these open-sourced or API-available models to achieve open-world visual perception remains an open question. 
In this paper, we introduce VisionGPT to consolidate and automate the integration of state-of-the-art foundation models, thereby facilitating vision-language understanding and the development of vision-oriented AI. 
VisionGPT builds upon a generalized multimodal framework that distinguishes itself through three key features: 
(1) utilizing LLMs (e.g., LLaMA-2) as the pivot to break down users' requests into detailed action proposals to call suitable foundation models; 
(2) integrating multi-source outputs from foundation models automatically and generating comprehensive responses for users;
(3) adaptable to a wide range of applications such as text-conditioned image understanding/generation/editing.
This paper outlines the architecture and capabilities of VisionGPT, demonstrating its potential to revolutionize the field of computer vision through enhanced efficiency, versatility, and generalization, and performance. 
Our code and models will be made publicly available.
\end{abstract}

\section{Introduction}
\label{sec:intro}

In the rapidly evolving era characterized by generative AI, large language models (LLMs) and vision foundation models have emerged as front runners in advancing the research in natural language processing (NLP), computer vision (CV), and their intersection\cite{min2023recent,wang2023internimage,li2023large,koh2023generating}. Although foundation models like GPT-4 have established a remarkable benchmark, developing an interactive agent to achieve open-world visual perception remains an ever-changing frontier brimming with application potential\cite{gsam2023,shen2023hugginggpt,wu2023visual}, e.g., security surveillance and autonomous driving. 

\begin{figure}[tb]
  \centering
  \begin{subfigure}[t]{0.32\linewidth}
    \includegraphics[width=\linewidth]{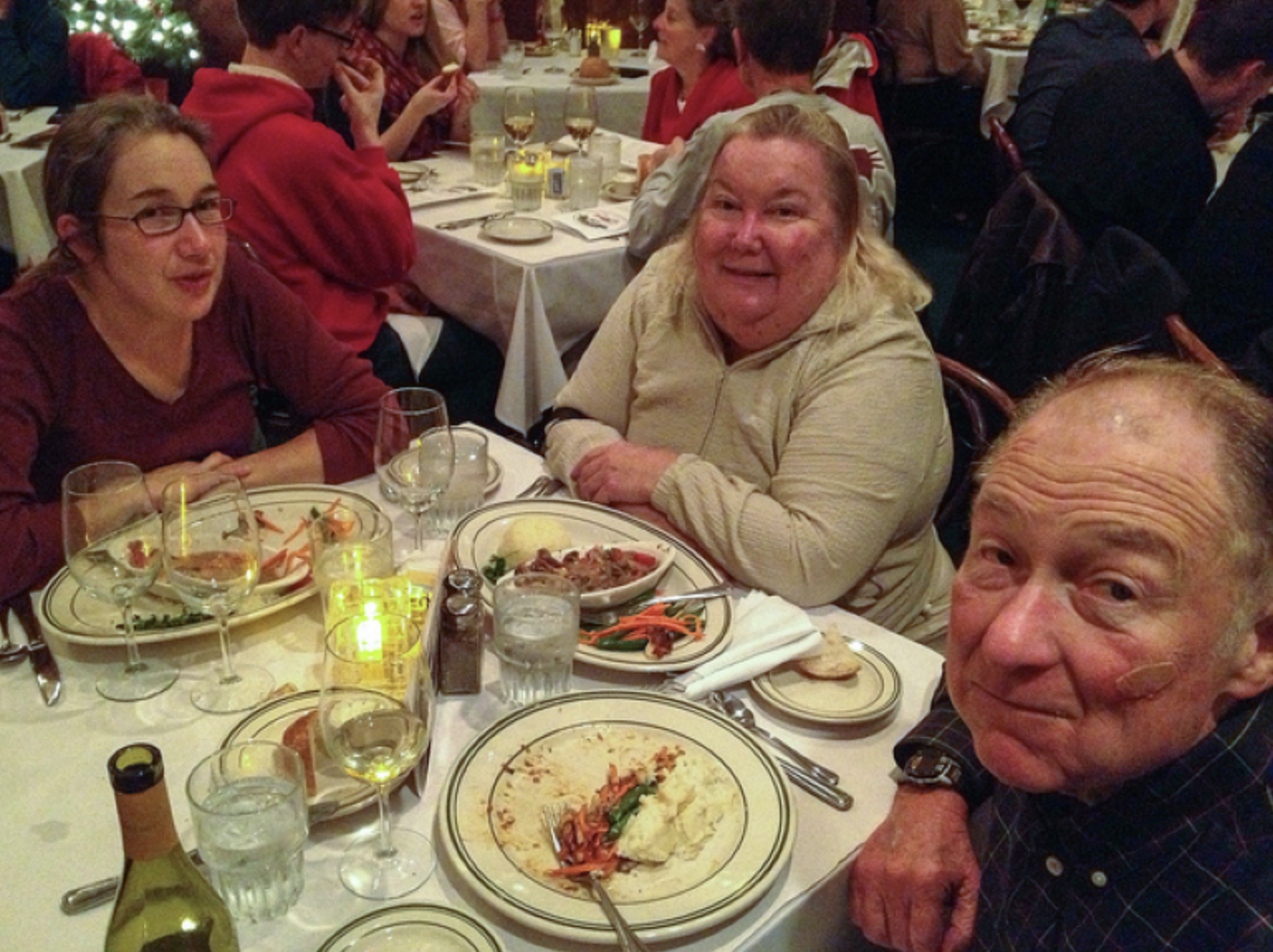}
    \caption{Without processing}
    \label{fig:intro_example_a}
  \end{subfigure}
  \hfill
  \begin{subfigure}[t]{0.32\linewidth}
    \includegraphics[width=\linewidth]{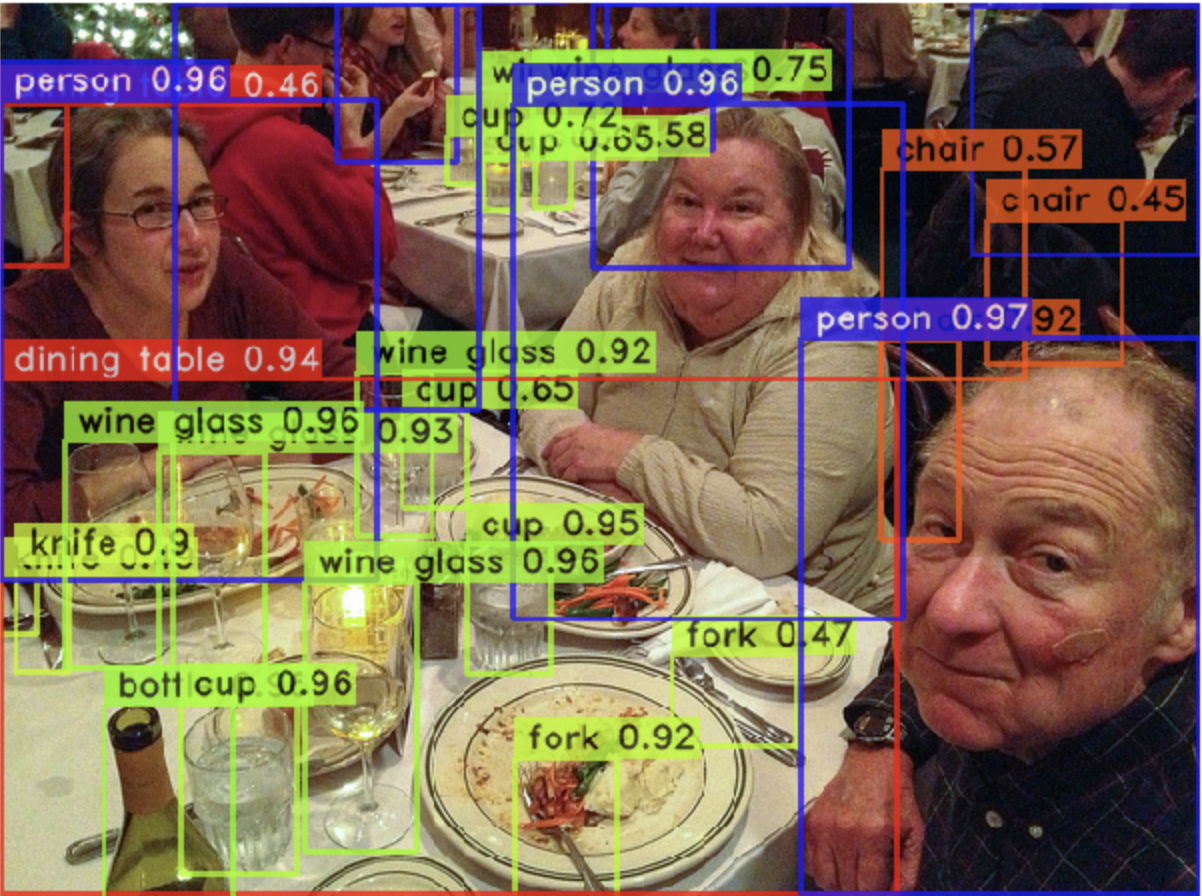}
    \caption{With YOLO detection}
    \label{fig:intro_example_b}
  \end{subfigure}
  \hfill
  \begin{subfigure}[t]{0.32\linewidth}
    \includegraphics[width=\linewidth]{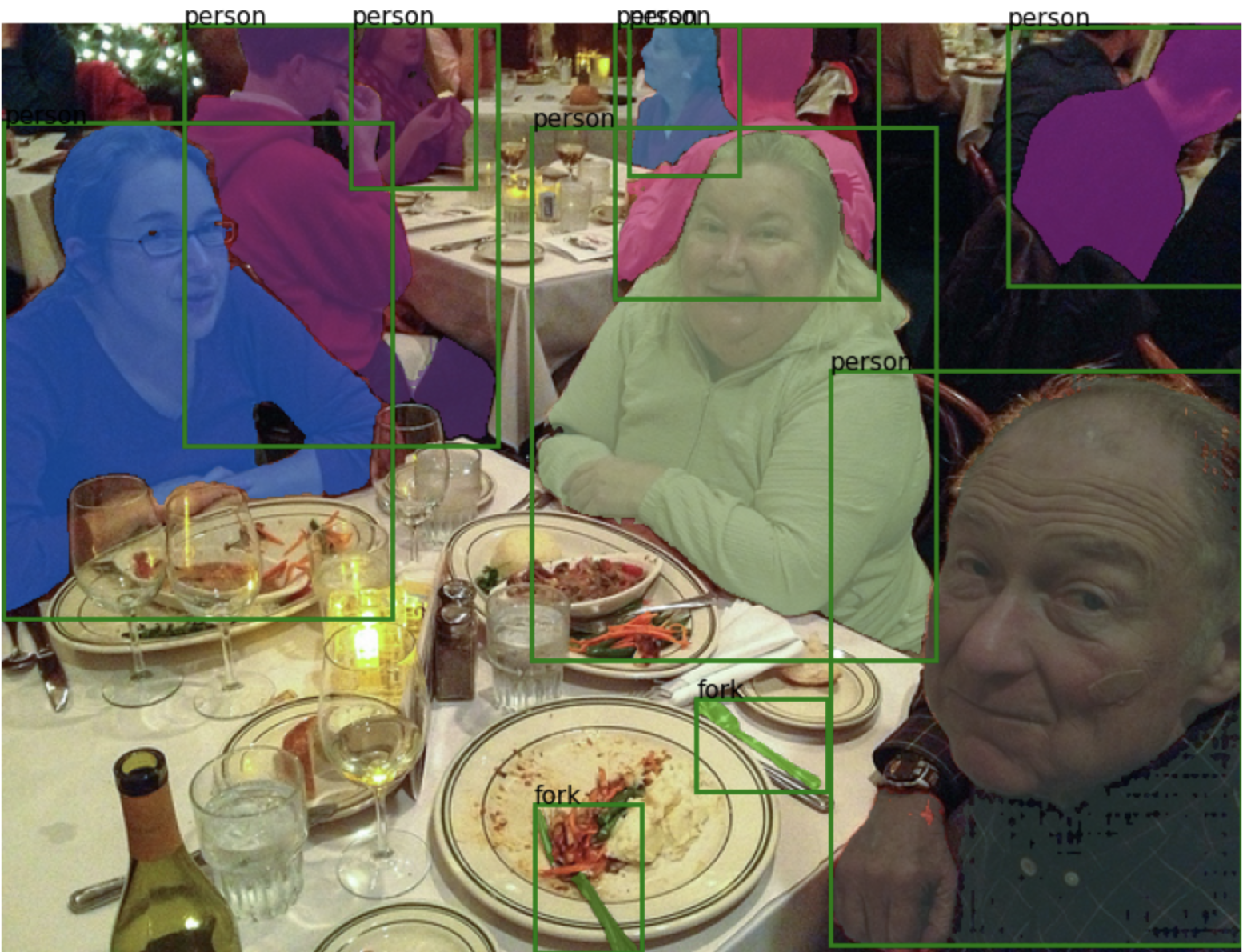}
    \caption{With YOLO detection first and then SAM segmentation}
    \label{fig:intro_example_c}
  \end{subfigure}
  \caption{An example of how YOLO and SAM collaborate to achieve instance segmentation. Given image (a) and interested objects ``fork'' and ``person'', YOLO is applied for detection, based on which SAM can solely segment regions of interest to improve efficiency.}
  \label{fig:intro_example}
\end{figure}

\begin{figure}[t]
    \centering
    \includegraphics[width=1\linewidth]{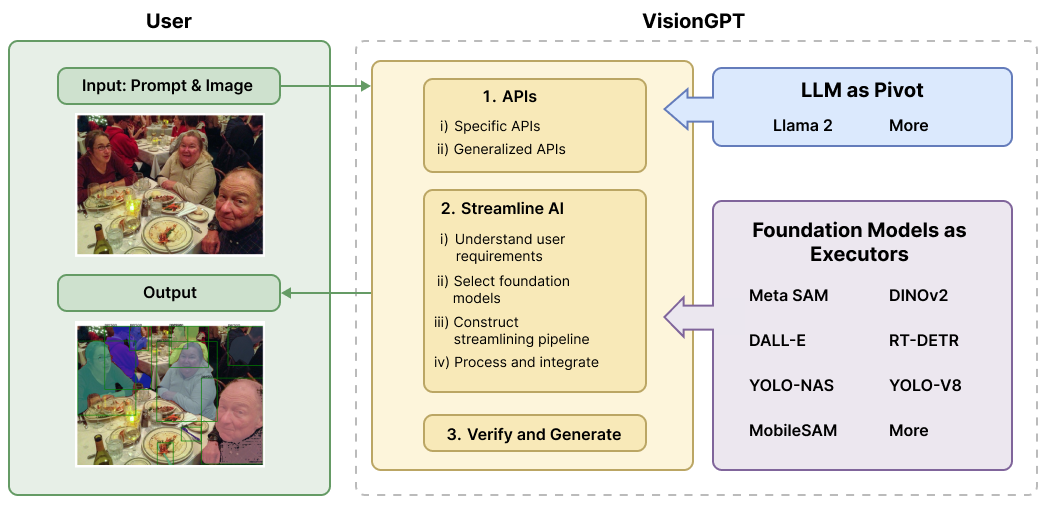}
    \caption{Overview of VisionGPT. It leverages LLMs (e.g., Llama 2\cite{llama2}) as the pivot to understand user intentions and state-of-the-art foundation models as executors to produce desired output. Building upon such a generalized multimodal framework, VisionGPT can accommodate any up-to-date foundation models and facilitate collaboration between models to address complex and challenging tasks.}
    \label{fig:visiongpt_overview}
\end{figure}

\begin{figure}[htbp]
    \centering
    \includegraphics[width=1\linewidth]{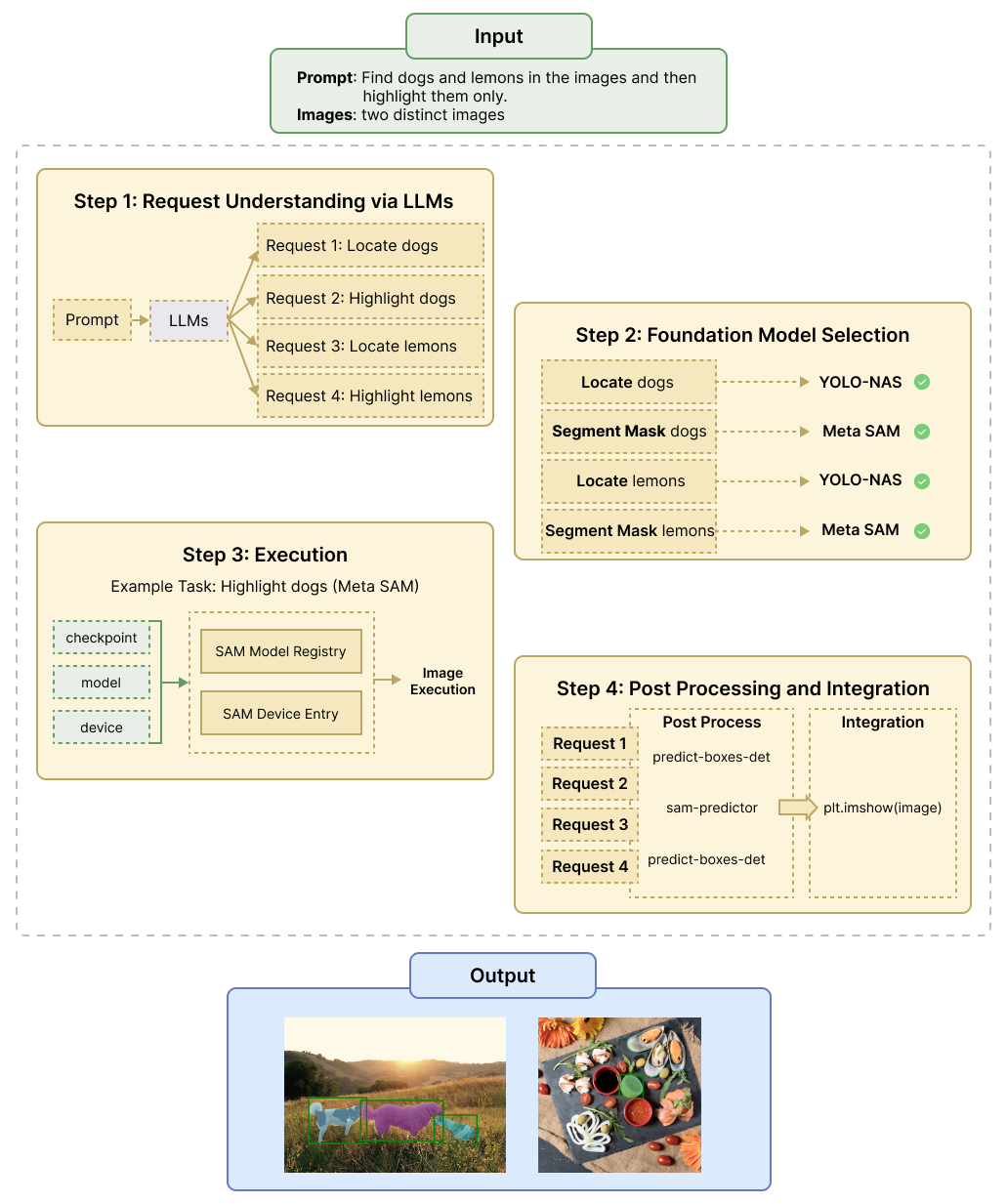}
    \caption{An example of the workflow of VisionGPT, which can be broken down into four key steps: 1) request understanding via LLMs, 2) foundation model selection,  3) execution, and 4) post processing and integration.} 
    \label{fig:visiongpt_example}
\end{figure}

In response to the demand for visual perception in open-world scenarios, researchers are actively exploring task-oriented vision foundation models, e.g., segment anything model\cite{kirillov2023segment} (SAM) and detect everything model\cite{wang2023detecting}. However, each foundation model has its distinct knowledge, making it hard to solve complex tasks with a single model. Instead, combining the intelligence and capacity of the off-the-shelf models turns out to be a more efficient and environment-friendly solution. Let's consider the challenging instance segmentation task, where distinct objects of the same class need to be differentiated from each other through different colored masks, and take YOLO\cite{Jocher_Ultralytics_YOLO_2023,supergradients} and SAM as examples. As depicted in \cref{fig:intro_example}, we can unite YOLO and SAM to achieve instance segmentation efficiently. Specifically, instead of indiscriminately segmenting all objects in \cref{fig:intro_example_a}, we can first apply YOLO to detect objects at a glance. Then given the interested objects ``fork'' and ``person'', we can retain some of the detection results shown in \cref{fig:intro_example_b}, based on which SAM can generate unique masks for the interested objects (\cref{fig:intro_example_c}). 
While the above example showcases the feasibility of model collaboration, the automation of ``divide and rule'' for various user requests and task scenarios remains unresolved.

In this paper, we introduce VisionGPT to enhance open-world visual perception from a systems-building view. VisionGPT operates on a multimodal framework that combines the strengths of state-of-the-art (SOTA) LLMs and vision foundation models, including Llama 2\cite{llama2}, SAM-series\cite{kirillov2023segment,faster,fast}, YOLO-series\cite{Jocher_Ultralytics_YOLO_2023,supergradients}, DINO\cite{dinov2,darcet}, Detectron2\cite{detectron2}, DALL-E\cite{zeroshot,dalle}, CLIP\cite{clip}, etc. \cref{fig:visiongpt_overview} illustrates the cores of VisionGPT: it utilizes LLM as pivot to understand the custom requests of users and automate the whole workflow whereas leverages different foundation models as executors to generate desired outputs. The workflow of our VisionGPT can be decomposed into a few steps: 
\begin{enumerate}
    \item \textbf{Request Understanding via LLMs:} In this step, the LLM interprets the user's request and breaks down the user input into detailed action proposals. The concomitant visual data is also uploaded to the whole system.
    \item \textbf{Foundation Model Selection:} Appropriate foundation models are selected depending on the action proposals.
    \item \textbf{Execution:} Foundation models act as computing engines and provide their expert insights to the uploaded visual data.
    \item \textbf{Post Processing and Integration:} The multi-source outputs from foundation models are first post-processed, then integrated, and finally returned to the user.
\end{enumerate}

To better understand the workflow of VisionGPT, we give an example in \cref{fig:visiongpt_example}. Specifically, consider two distinct images rather than a single image as input. A user might make the following request: \textit{“Find dogs and lemons in the images and then highlight them only”}. VisionGPT will have a LLM interpret this request and break it down into several reasonable action proposals as below:
\begin{enumerate}[label=\Alph*.]
    \item Locate dogs
    \item Segment dogs
    \item Locate lemons
    \item Segment lemons
    \item Integrate all results (optional)
\end{enumerate}
After the user request has been interpreted, the best-suited foundation model will be selected. In this case, action proposal A calls for a simple object detection of all dogs, so models like YOLO will be called upon. Action proposal B asks for the dogs to be highlighted, which means that the segmentation abilities of SAM will be needed. As mentioned earlier, SAM can leverage the detection results of dogs to reduce inference latency. Action proposals C and D will repeat as the above. 

It is noteworthy that VisionGPT is not limited to the models listed in \cref{fig:visiongpt_overview}, i.e., VisionGPT is flexible to accommodate up-to-date foundation models for better task-solving skills, making it adaptable to a wide range of applications such as text-conditioned image understanding/generation/editing. Moreover, VisionGPT serves multiple purposes that align with the current and future needs of the AI community. By providing a generalized multimodal framework, this project will help accelerate the development of vision-oriented AI and bridge the gap between the status quo of LLMs and the emergent CV multimodal paradigm. Last but not least, this paper will explore the capabilities, methodologies, and potential future applications of this technology.

\section{Related Works}
\label{sec:related_works}

The integration of state-of-the-art (SOTA) computer vision (CV) technologies and large language models (LLMs) has become increasingly prevalent in the field of artificial intelligence. VisionGPT uniquely contributes to this domain by synergizing SOTA vision models, such as YOLO \cite{yolo} and Meta SAM \cite{kirillov2023segment}, into a comprehensive multimodal framework. 

In terms of bridging frameworks with LLMs, HuggingGPT \cite{hugginggpt} emerges as a pertinent reference, connecting the extensive model repository HuggingFace \cite{huggingface} to LLMs like ChatGPT. Although this approach aligns with VisionGPT in facilitating interactions with LLMs, VisionGPT distinguishes itself by prioritizing a seamless integration and automation of SOTA vision models, thereby fostering a robust and vision-focused AI ecosystem.

Grounded SAM \cite{gsam2023} extends the capabilities of Meta SAM through training with language instructions, enhancing its object segmentation proficiency. This innovation resonates with VisionGPT’s objective of harnessing and amplifying the unique capabilities of existing models to advance vision-oriented AI.

MiniGPT-4 \cite{minigpt} represents another stride towards integrating visual and linguistic modalities, aligning a visual encoder with the advanced LLM, Vicuna. Although there are parallels in modality integration between MiniGPT-4 and VisionGPT, VisionGPT stands out by offering a holistic integration and automation of various SOTA vision models, ensuring versatility and peak performance across diverse applications.

The work on VoxPoser \cite{voxposer} serves as a related endeavor to VisionGPT in integrating Large Language Models (LLMs) with robotic manipulation tasks. VoxPoser uniquely synthesizes robot trajectories and constructs 3D value maps based on natural language instructions, highlighting its strength in dealing with a wide array of manipulation tasks. VisionGPT, on the other hand, extends the application of LLMs beyond manipulation, aiming to create a generalized and automated framework for various vision-oriented tasks. Although they share common ground in leveraging LLMs for robotic methods, each framework carves out its niche, contributing uniquely to the intersection of language models and vision-based robotic tasks.

In "Physically Grounded Vision-Language Models for Robotic Manipulation" \cite{pgvlm}, the authors address the limitations of current Vision-Language Models (VLMs) in understanding physical concepts crucial for robotic manipulation. They introduce a new dataset, PhysObjects, to enhance the VLM’s comprehension of these concepts, resulting in improved robotic planning performance. While VisionGPT focuses on creating a unified and automated framework for a variety of vision-based tasks using Large Language Models, \cite{pgvlm} emphasizes physically-grounding VLMs to augment their utility in robotic manipulation tasks. Both works underscore the significance of integrating language models with visual perception for robotic applications, albeit with different focuses and methodologies.

Foundation models and GPTs grow so fast. There are other related works, including Visual ChatGPT, WorldGPT, Flamingo, VLMo, VIOLET, and more \cite{visual,worldgpt,flamingo,vlmo,violet}. They mainly focused on ChatGPT integration or transformer-based solutions. VisionGPT differentiates them with its generalized multimodal framework. Its own framework can fine-tune the LLM \cite{llama} for its specific streamlining purpose \cite{llmzs}.

In conclusion, despite the presence of related works in CV and LLM integration, VisionGPT still offers unique capabilities. It does so by unifying and automating the capabilities of SOTA vision models, optimizing multimodal interactions, and delivering a streamlined and efficient user experience.

\section{VisionGPT}
\label{sec:framework}

\subsection{Overview}
VisionGPT operates as a cooperative agent designed to facilitate vision-language understanding and vision-oriented AI, reinventing our prior work\cite{unifiedvisiongpt}. VisionGPT combines the strengths of large language models (LLMs) and vision foundation models, adopting an in-context learning approach to generalize and automate a variety of complex tasks based on natural language instructions. 
When confronted with a user's request, VisionGPT initiates an automated deployment of the complete workflow. 
This orchestrates the collaboration and utilization of expert models, such as the YOLO model and the SAM model, to successfully accomplish the designated objectives set up by the user. 

Establishing a connection between LLMs and vision foundation models, VisionGPT opens up a realm of possibilities that could potentially lead to the ultimate customization of user requests. 
This synergy among foundation models creates a dynamic ecosystem where the linguistic and reasoning capabilities of LLMs can be seamlessly integrated with the image processing and understanding capability of vision foundation models. 
When a user submits a request, VisionGPT first utilizes LLMs to parse the natural language input to extract a number of details, context, and intent and simultaneously analyzes any accompanying images or visual data linked to the request. 
Then, VisionGPT arranges collaboration among foundation models, effectively translating the user's request into scene understanding and object analysis within images or visual data. 
This connection paves the way for a highly adaptive and context-aware system, capable of adapting its responses to the user's preferences, language nuances, and the specific content of the visual data provided. 
As the connection between VisionGPT and future LLMs evolves, the potential for ultimate customization of user requests becomes increasingly tangible, such as performing specific tasks relating to object recognition, scene understanding, or even creative image generation, tailored to the user's unique needs.

VisionGPT has four main components: APIs, Streamline AI, Verify and Generate, and Fine Tuning. Please refer to \cref{fig:visiongpt_overview} for the first 3 components.
\begin{enumerate}
    \item \textbf{APIs:} VisionGPT supports two types of APIs: specific APIs and generalized APIs. The specific APIs can be some simple or standard APIs for some common vision AI operations, for example, $\mathtt{labelObjects}$($<\mathtt{object}$ $\mathtt{name}>$, $<\mathtt{image}$ $\mathtt{location}>$). The generalized APIs are flexible for inputs: a text prompt for instruction and a list of images or videos. The prompt can control and instruct the operations on images or videos.
    \item \textbf{Streamline AI:} VisionGPT has intelligent logic to automate the process of vision-language understanding based on LLMs and the integrated SOTA vision foundation models.
    \item \textbf{Verify and Generate:} VisionGPT is unique to verify the results against the inputs for the best results. It will retry if it detects something wrong. For example, it can use vision-text similarity measured by CLIP\cite{clip} as a proxy to decide whether to rerun. The generation step is an additional step to use other vision tools, such as OpenCV and GAI (Generative AI) to generate based on the API requirements or the input prompts. 
    \item \textbf{Fine Tune:} Although the LLM used in our VisionGPT, i.e., Llama 2, is one of the best open-source LLMs, it still has limits in the specific domains and streamlining generalization. So VisionGPT has its dedicated database and historical vision-related dataset to fine-tune the Llama 2 for a better domain-specific LLM. 
\end{enumerate}

VisionGPT is an open framework from the APIs to the internal streamlining logic. The goal of VisionGPT aims to provide a generalized multimodal framework for streamlining versatile vision AI.

\subsection{Problem Formulation}
Consider a scenario where we are given a natural language instruction $L$ that describes a specific task. The goal of VisionGPT is to interpret $L$ and translate it into a set of action proposals $A=\{a_1, a_2, \dots, a_N\}$, where each element represents an individual operation such as object recognition, image segmentation, or feature extraction. The core challenge is to ensure that the interpretation of $L$ accurately reflects the intended operations, aligning the output with the user's expectations. The problem can be mathematically formulated as:
\begin{equation}
\label{eq:formulation1}
    \min_{\mathcal{L}} \mathcal{L}(f(L), A) + \mathcal{R}(A),
\end{equation}
where $f(\cdot)$ denotes the inference process of the LLM (i.e., Llama 2 in this work), $f(L)$ the generated action proposals, $\mathcal{L}(\cdot)$ the loss function measuring the discrepancy between the generated and ground-truth action proposals, and $\mathcal{R}(A)$ the regularization term ensuring the action proposals are well-defined and executable. Given that it could be difficult to minimize $\mathcal{L}(f(L), A)$ without downstream fine-tuning, we further introduce a piece of system information $S$ to steer the LLM to generate desired outputs. Thus, \cref{eq:formulation1} can be modified to
\begin{equation}
\label{eq:formulation2}
    \min_{\mathcal{L}} \mathcal{L}(f(\{S; L\}), A) + \mathcal{R}(A),
\end{equation}
where $\{;\}$ denotes the text concatenation operation.

\subsection{Action Proposal Generation via In-Context Learning}
As indicated by \cref{eq:formulation2}, VisionGPT utilizes an advanced LLM to generate a corresponding set of action proposals ($A$) based on a natural language instruction ($L$) and a piece of system information ($S$). Motivated by the findings of previous works~\cite{brown2020language,wei2022chain,liu2023pre} that LLMs' performance can be improved by in-context learning, we implement $S$ as the demonstration of several paired $(L, A)$ examples to help the LLM understand the task format of action proposal generation. One example is given in \cref{tab:llm_input}. In-context learning ensures that the generated action proposals are better grounded in the linguistic context provided by the user, thus facilitating the accuracy of action proposal generation.

\begin{table}[t]
    \centering
    \caption{An example of the LLM input for action proposal generation.}
    \label{tab:llm_input}
    \begin{tabular}{l}
    \toprule
    LLM Input (\colorbox{gray!15}{System Information} ($S$) \& \colorbox{yellow!15}{User Input} ($L$))\\
    \midrule
    \rowcolor{gray!15}
    Please generate action proposals based on the following operation candidates: 
    \\
    \rowcolor{gray!15}
    "locate", "segment", .... The output should be as concise as possible.
    \\
    \rowcolor{gray!15}\\
    \rowcolor{gray!15}
    Input: highlight dogs and frogs in the image
    \\
    \rowcolor{gray!15}
    Output: "locate" dogs; "segment" dogs; "locate" frogs, "segment" frogs;
    \\
    \rowcolor{gray!15}\\
    \rowcolor{gray!15}...
    \\
    \rowcolor{gray!15}\\
    \rowcolor{gray!15}
    Input: \colorbox{yellow!15}{[User Input]} 
    \\
    \rowcolor{gray!15}Output: 
    \\
    \bottomrule
    \end{tabular}
\end{table}

\subsection{Task Automation via Few-Shot Generalization}
VisionGPT is designed to generalize across a wide array of tasks without relying on excessive task-specific training data. The recent work \cite{mishra2022cross} indicates that pre-trained language models can benefit from task-specific instructions when evaluated in terms of generalization to unseen tasks. Such cross-task generalization capacity ensures that LLM-based VisionGPT can automate action proposal generation across various contexts and applications based on only a few task-specific in-context examples, showcasing impressive efficiency and adaptability. 

\subsection{Joint Optimization for Coherent Task Execution}
To guarantee that the generated action proposals are not only linguistically aligned but also lead to successful execution, VisionGPT employs a joint optimization strategy. This approach considers both the semantic congruence between $L$ and $A$ and the practical feasibility of the action proposals. Through this comprehensive optimization, VisionGPT ensures coherent and effective task execution, aligning responses with the user's intent. 


\begin{figure}[htbp]
  \centering
  \begin{subfigure}[t]{0.65\linewidth}
      \begin{subfigure}[t]{0.49\linewidth}
        \includegraphics[width=\linewidth]{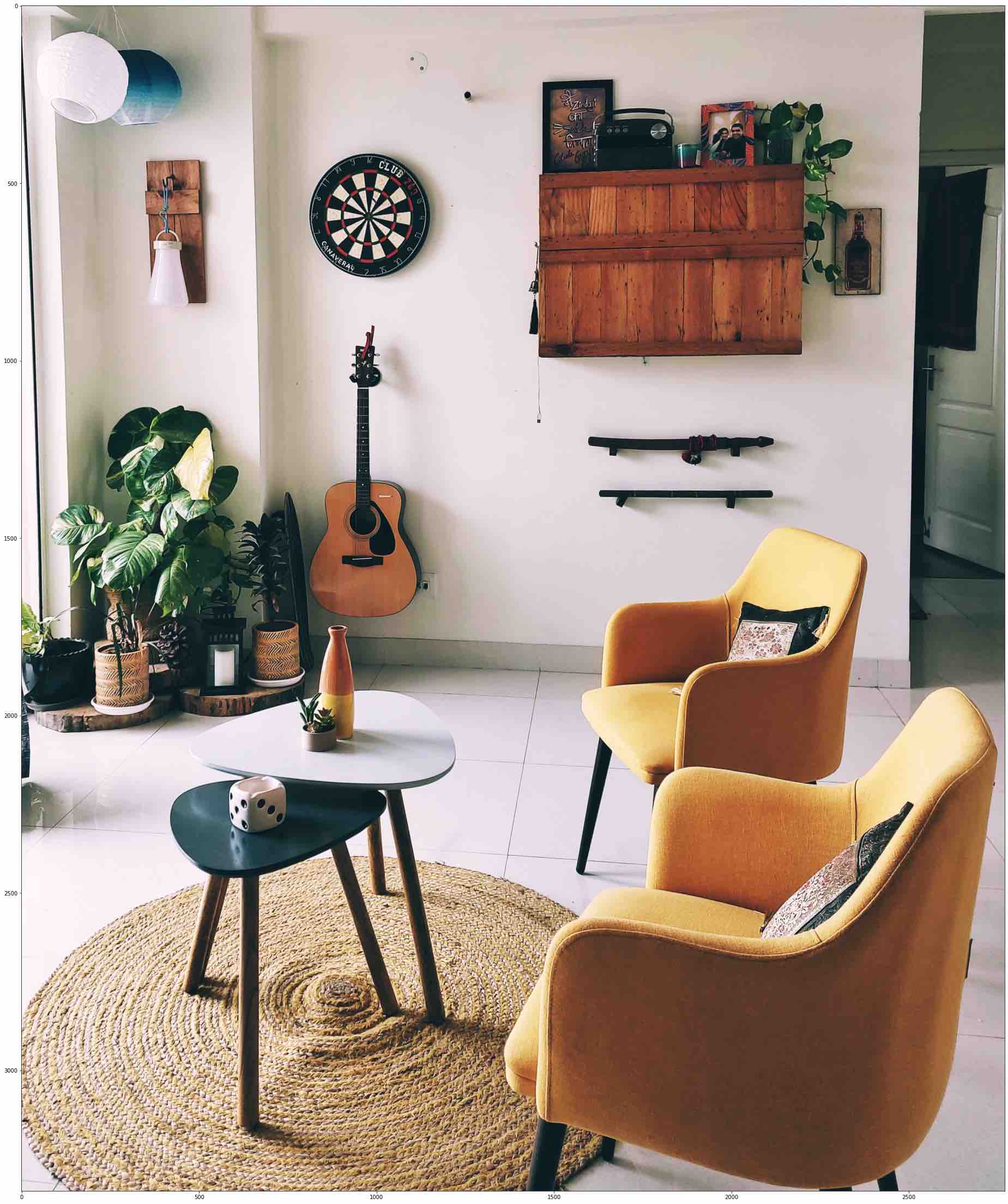}
      \end{subfigure}
      \hfill
      \begin{subfigure}[t]{0.49\linewidth}
        \includegraphics[width=\linewidth]{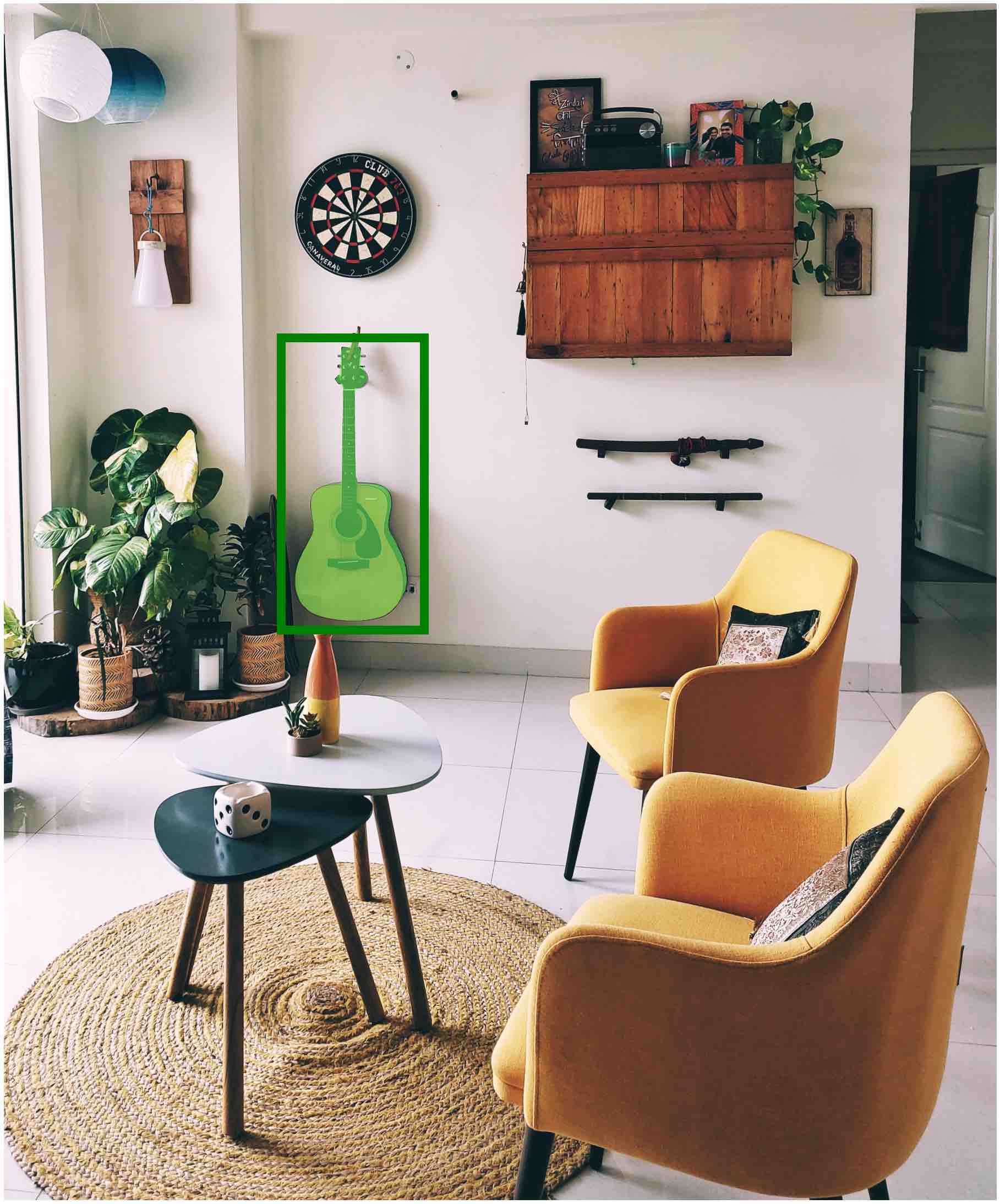}
      \end{subfigure}
      \caption{Find the guitar and segment it}
      \label{fig:cases_1}
  \end{subfigure}
  \\
  \begin{subfigure}[t]{0.65\linewidth}
      \begin{subfigure}[t]{0.49\linewidth}
        \includegraphics[width=\linewidth]{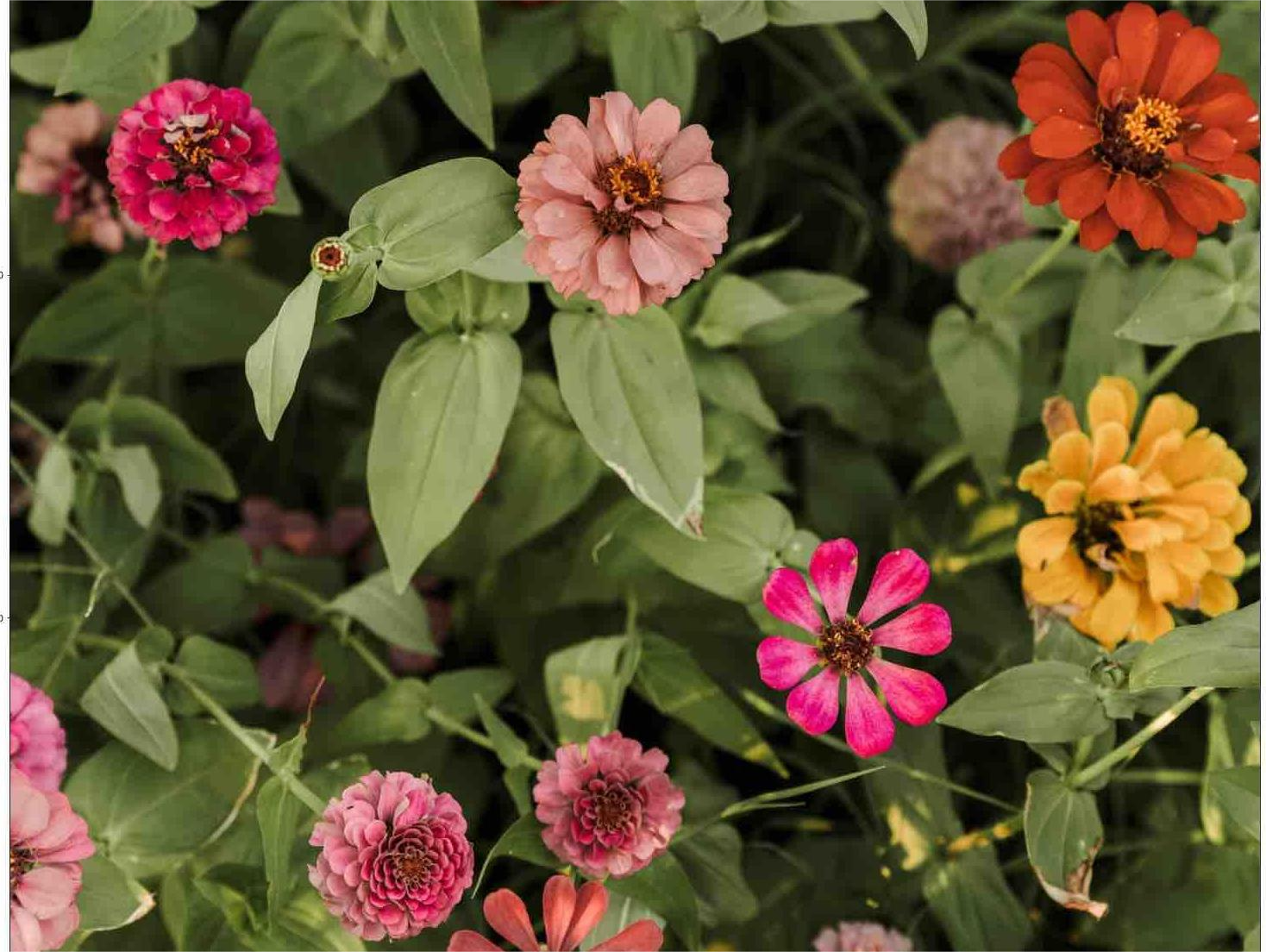}
      \end{subfigure}
      \hfill
      \begin{subfigure}[t]{0.49\linewidth}
        \includegraphics[width=\linewidth]{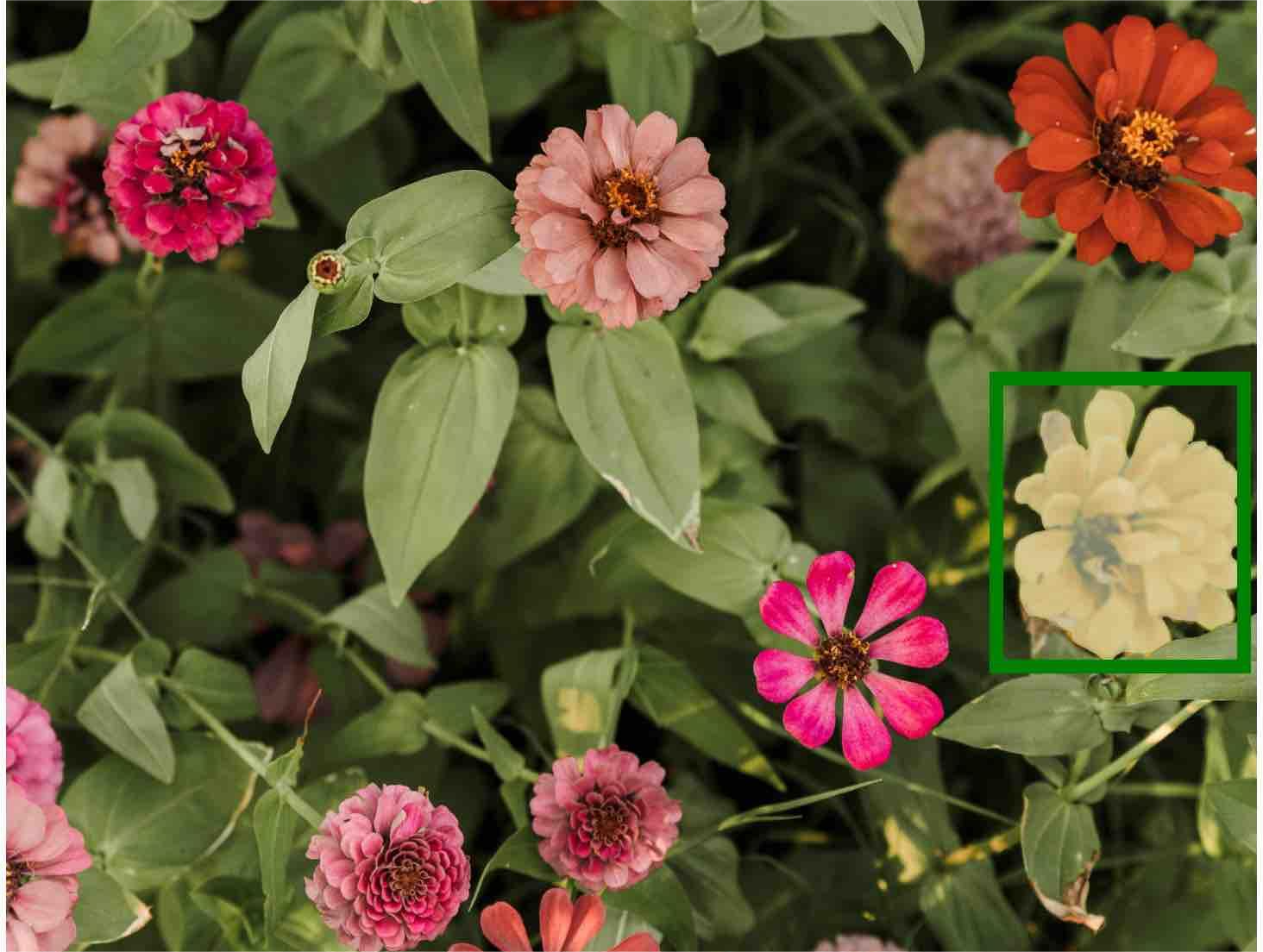}
      \end{subfigure}
      \caption{Find the yellow flower and segment it}
      \label{fig:cases_2}
  \end{subfigure}
  \\
  \begin{subfigure}[t]{0.65\linewidth}
      \begin{subfigure}[t]{0.49\linewidth}
        \includegraphics[width=\linewidth]{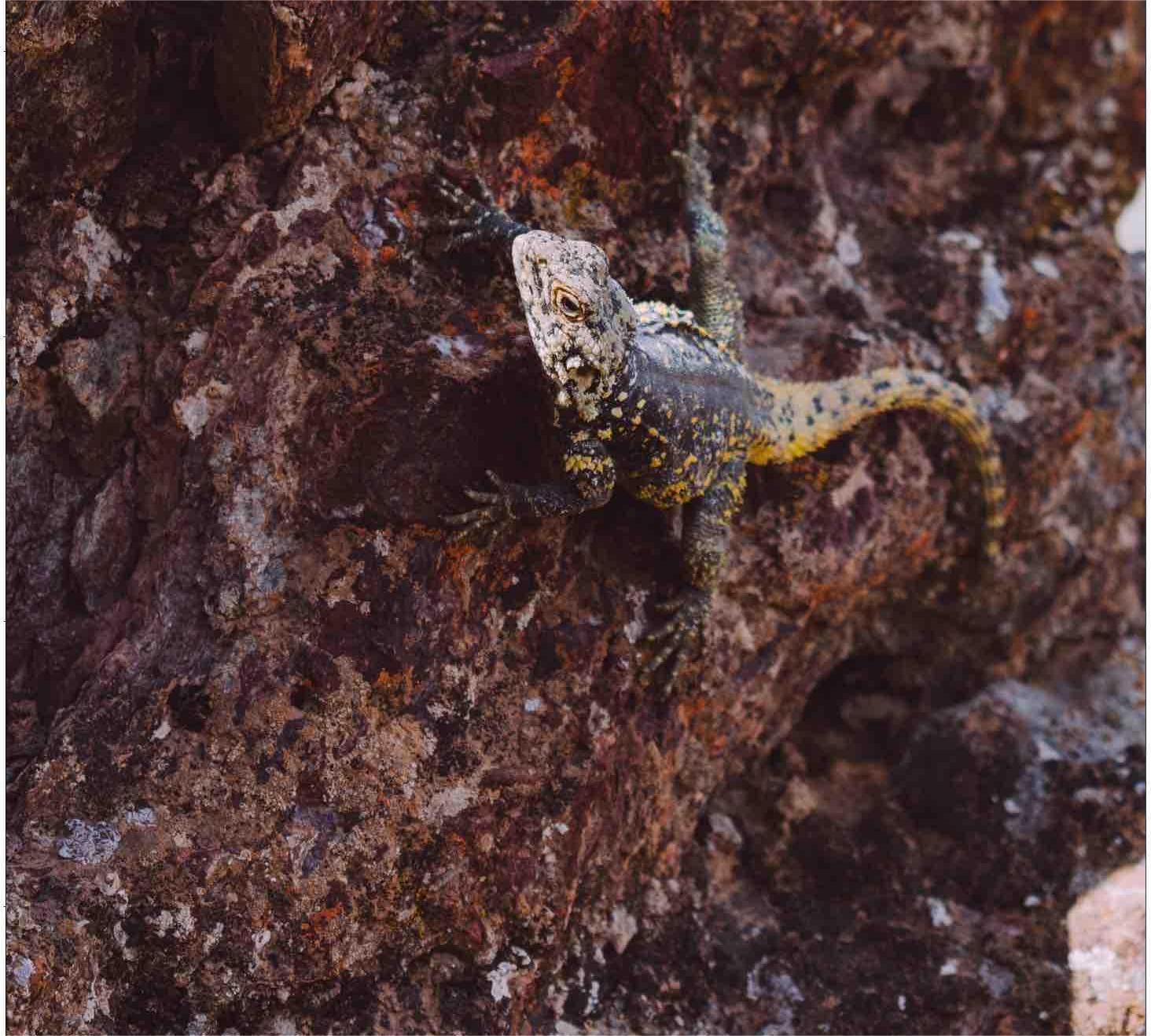}
      \end{subfigure}
      \hfill
      \begin{subfigure}[t]{0.49\linewidth}
        \includegraphics[width=\linewidth]{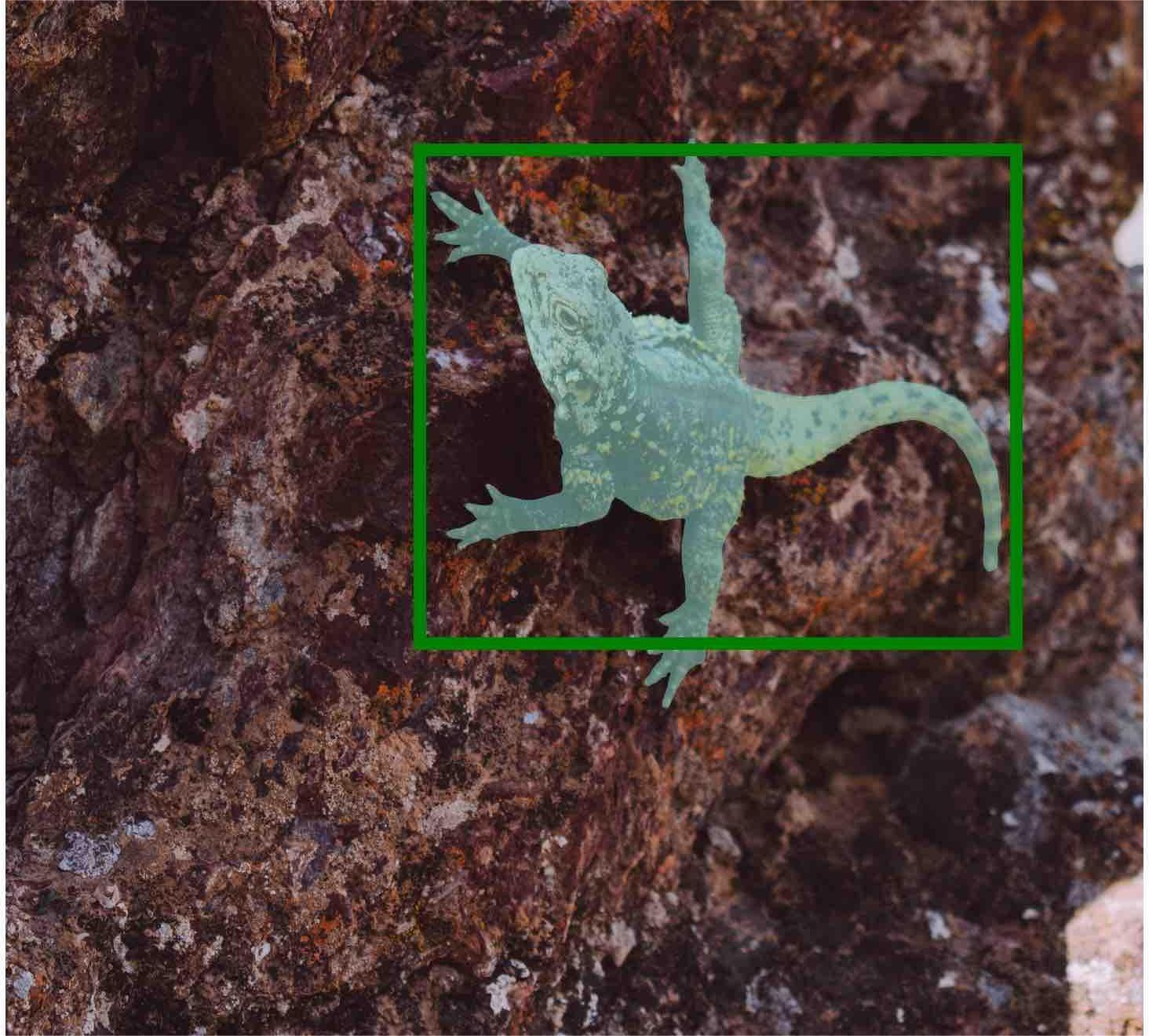}
      \end{subfigure}
      \caption{Find an animal and mask it}
      \label{fig:cases_3}
  \end{subfigure}
  \\
  \begin{subfigure}[t]{0.65\linewidth}
      \begin{subfigure}[t]{0.49\linewidth}
        \includegraphics[width=\linewidth]{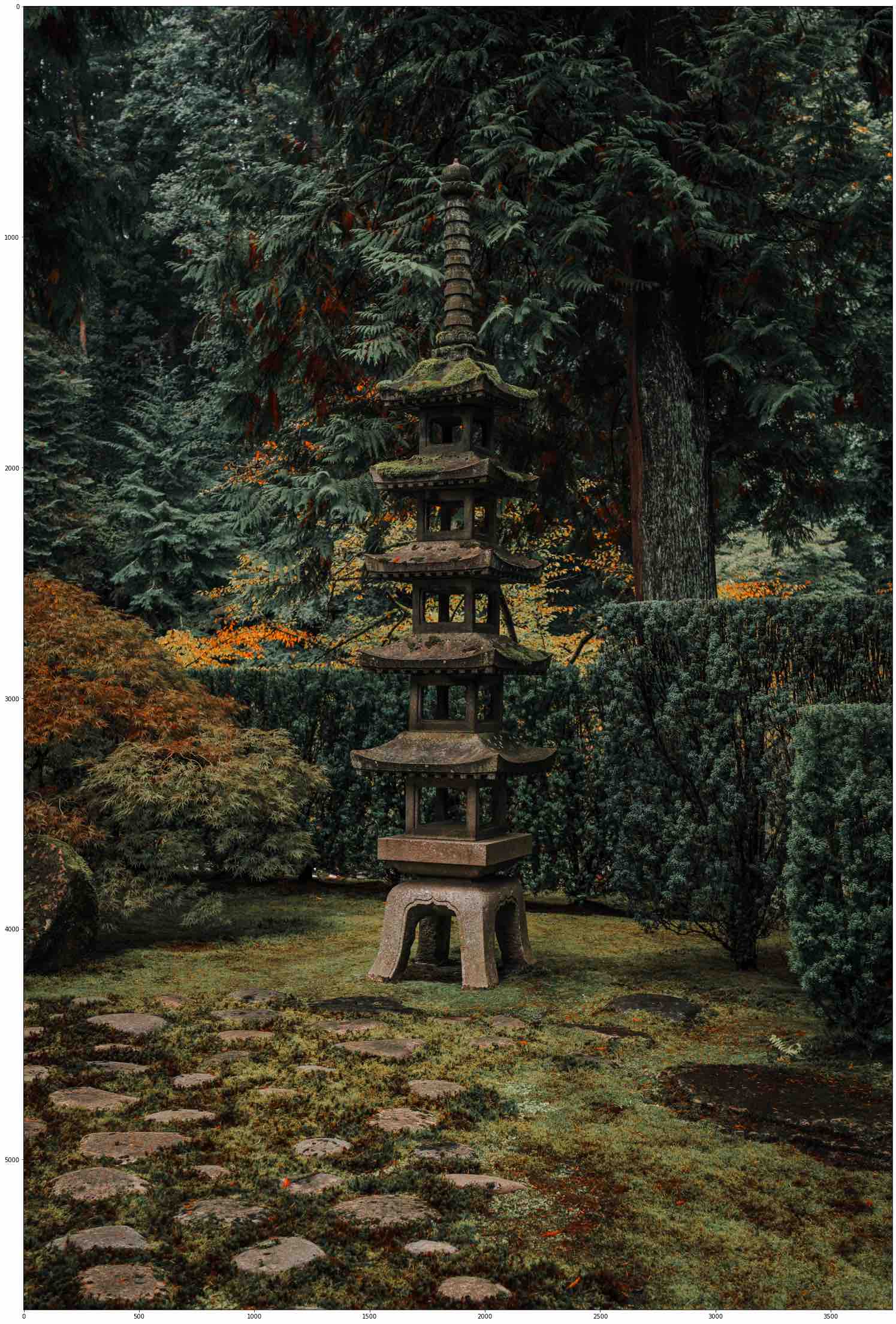}
      \end{subfigure}
      \hfill
      \begin{subfigure}[t]{0.49\linewidth}
        \includegraphics[width=\linewidth]{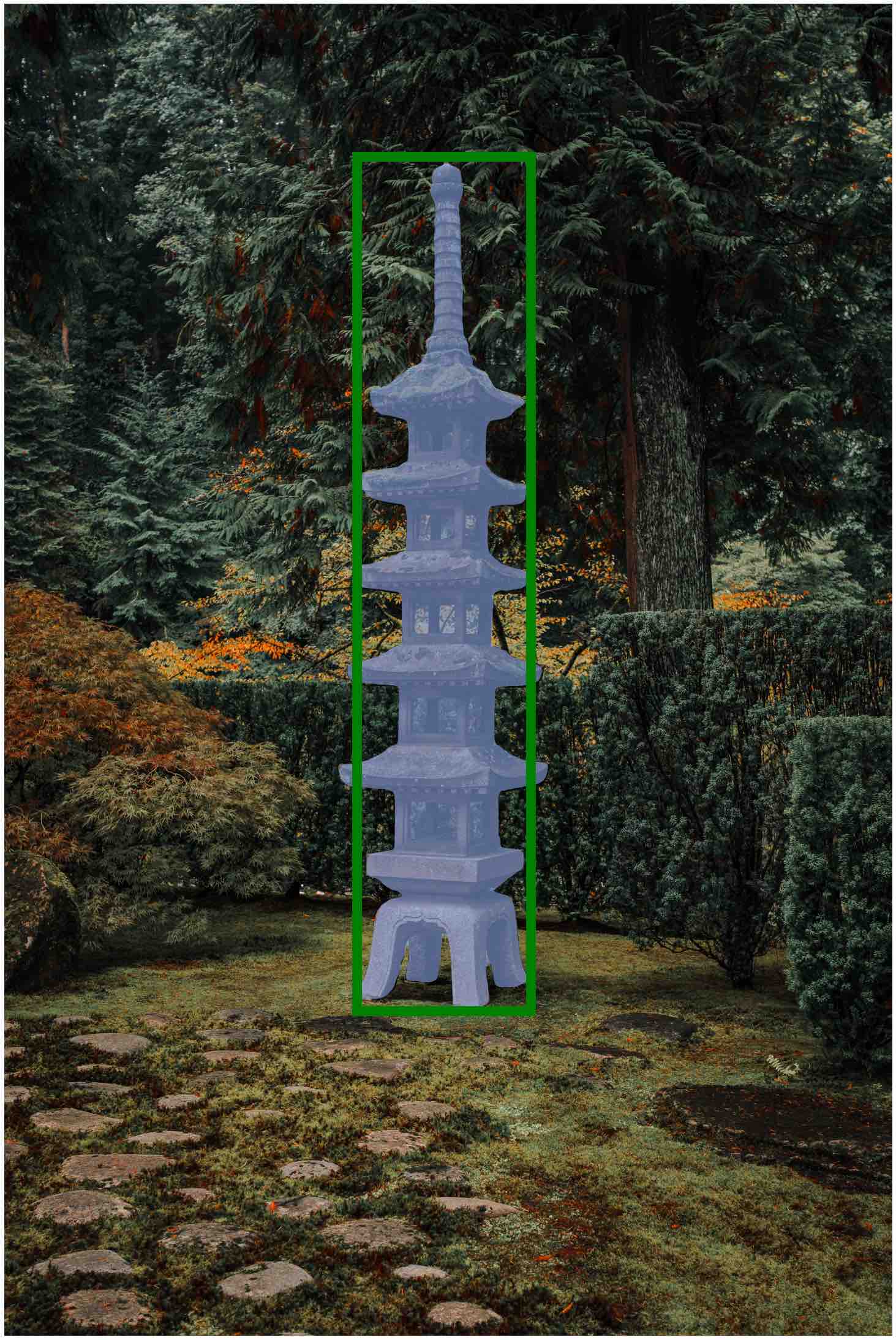}
      \end{subfigure}
      \caption{Mask any building in the image}
      \label{fig:cases_4}
  \end{subfigure}
\end{figure}

\begin{figure}[htbp]
  \ContinuedFloat
  \centering
  \begin{subfigure}[t]{0.65\linewidth}
      \begin{subfigure}[t]{0.49\linewidth}
        \includegraphics[width=\linewidth]{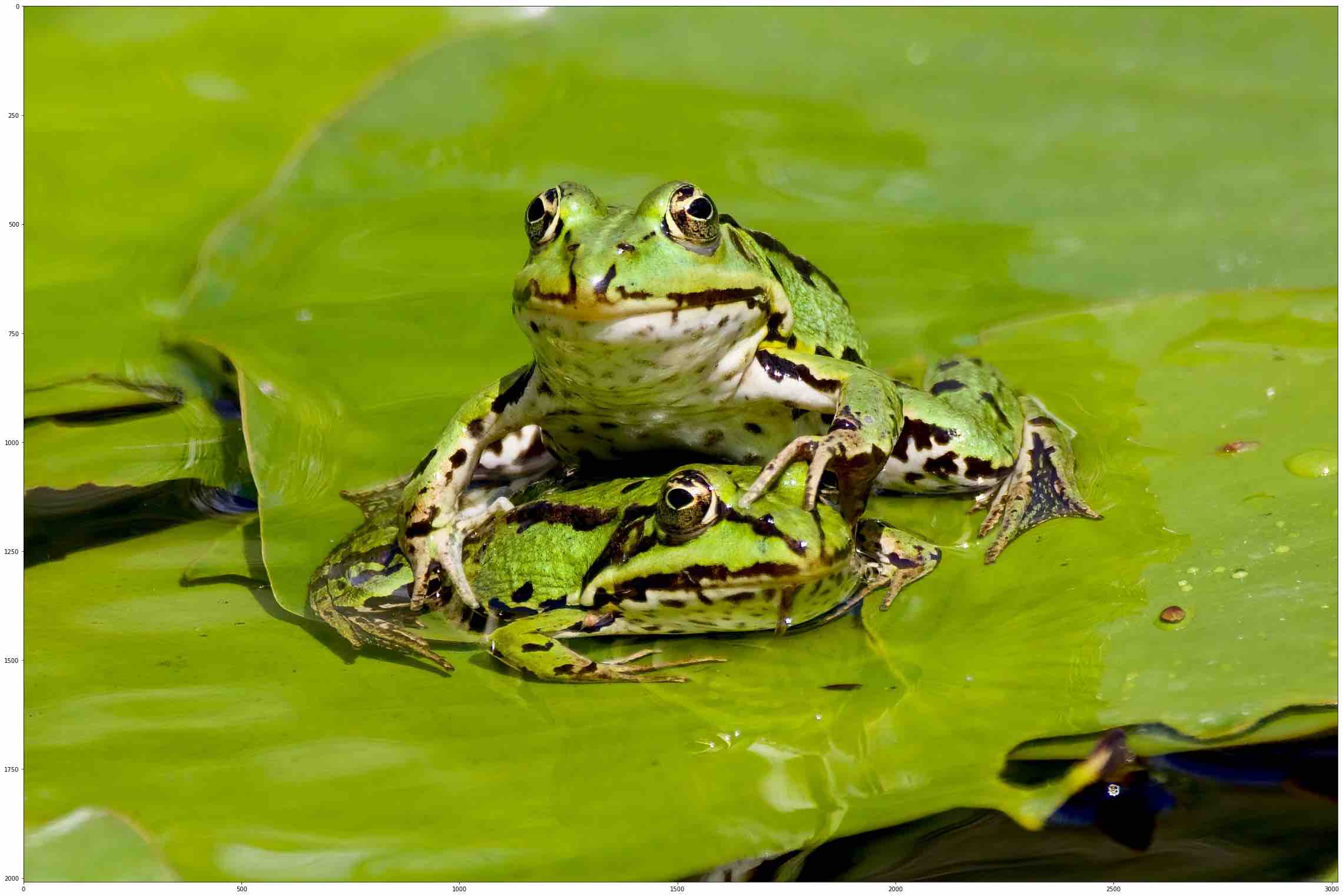}
      \end{subfigure}
      \hfill
      \begin{subfigure}[t]{0.49\linewidth}
        \includegraphics[width=\linewidth]{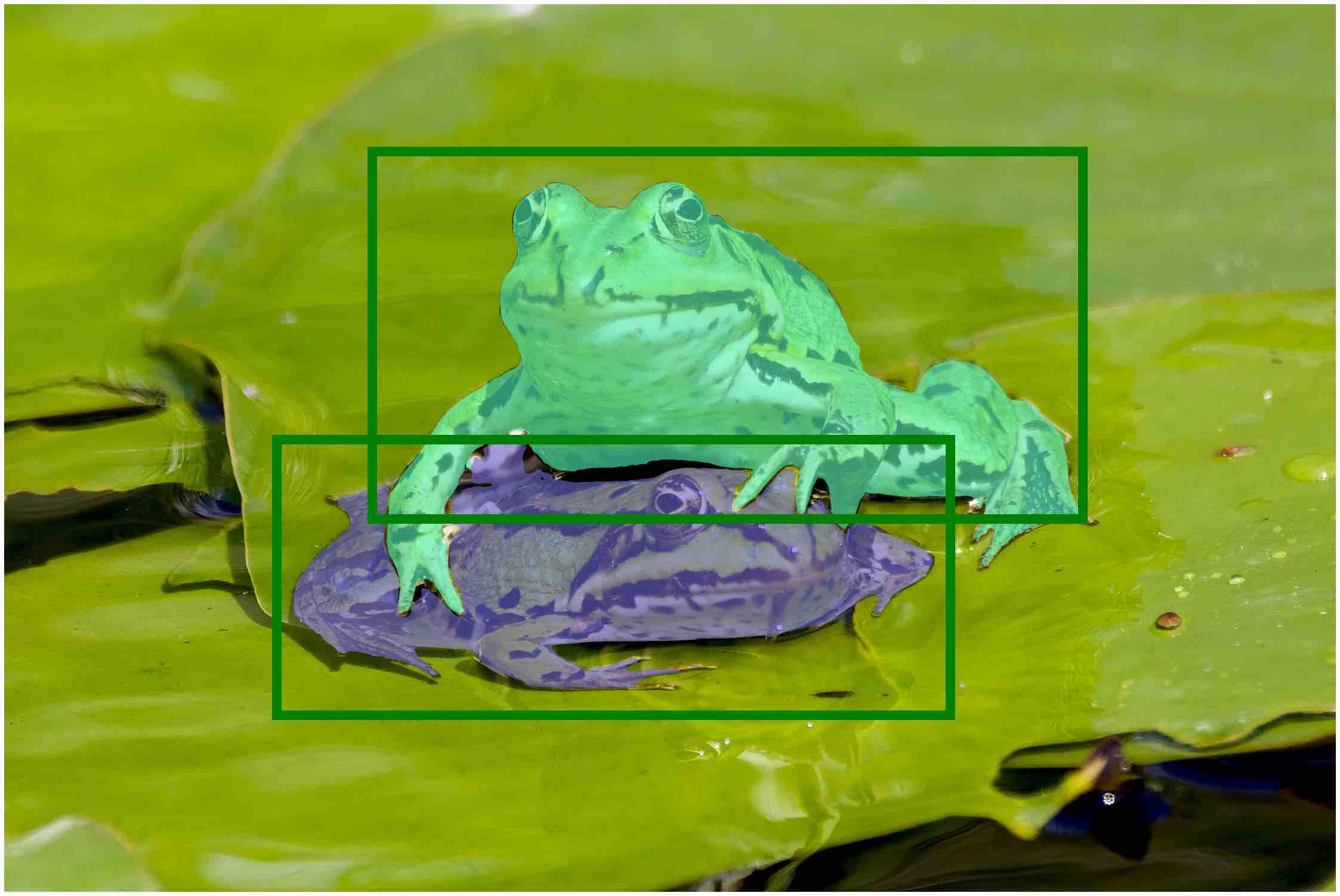}
      \end{subfigure}
      \caption{Highlight all frogs by masking them}
      \label{fig:cases_5}
  \end{subfigure}
  \\
  \begin{subfigure}[t]{0.65\linewidth}
      \begin{subfigure}[t]{0.49\linewidth}
        \includegraphics[width=\linewidth]{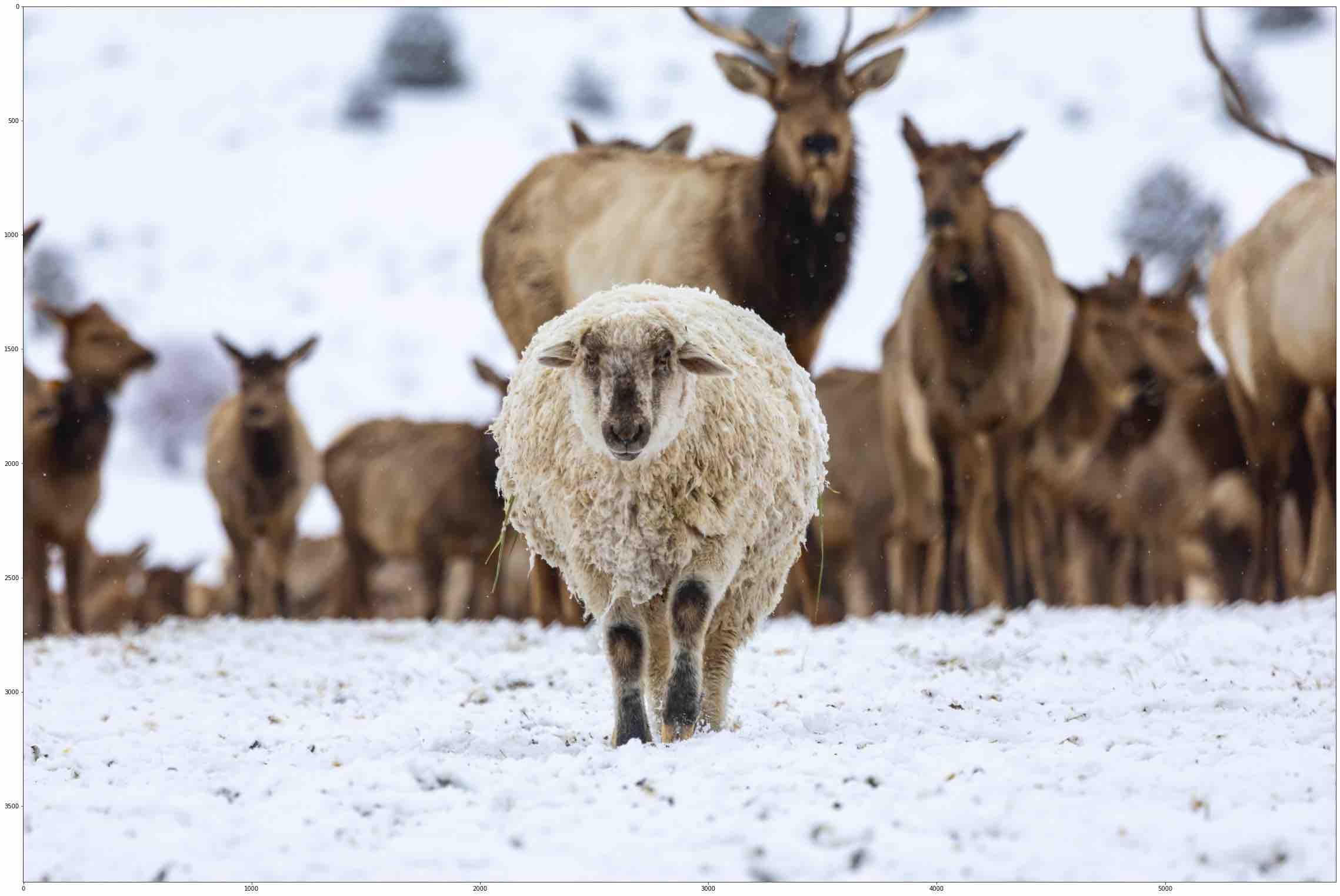}
      \end{subfigure}
      \hfill
      \begin{subfigure}[t]{0.49\linewidth}
        \includegraphics[width=\linewidth]{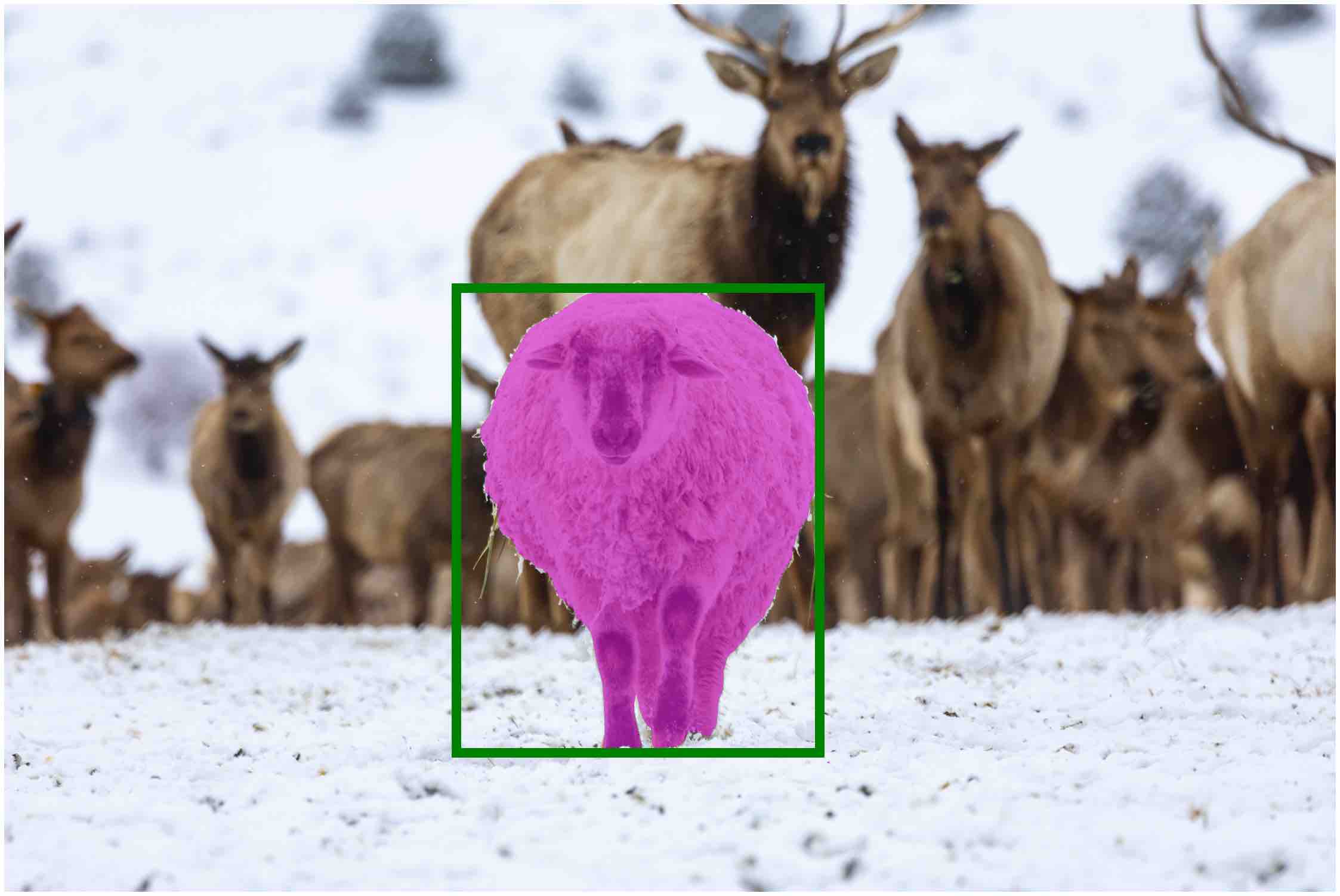}
      \end{subfigure}
      \caption{Find a different animal and segment it}
      \label{fig:cases_6}
  \end{subfigure}
  \\
  \begin{subfigure}[t]{0.65\linewidth}
      \begin{subfigure}[t]{0.49\linewidth}
        \includegraphics[width=\linewidth]{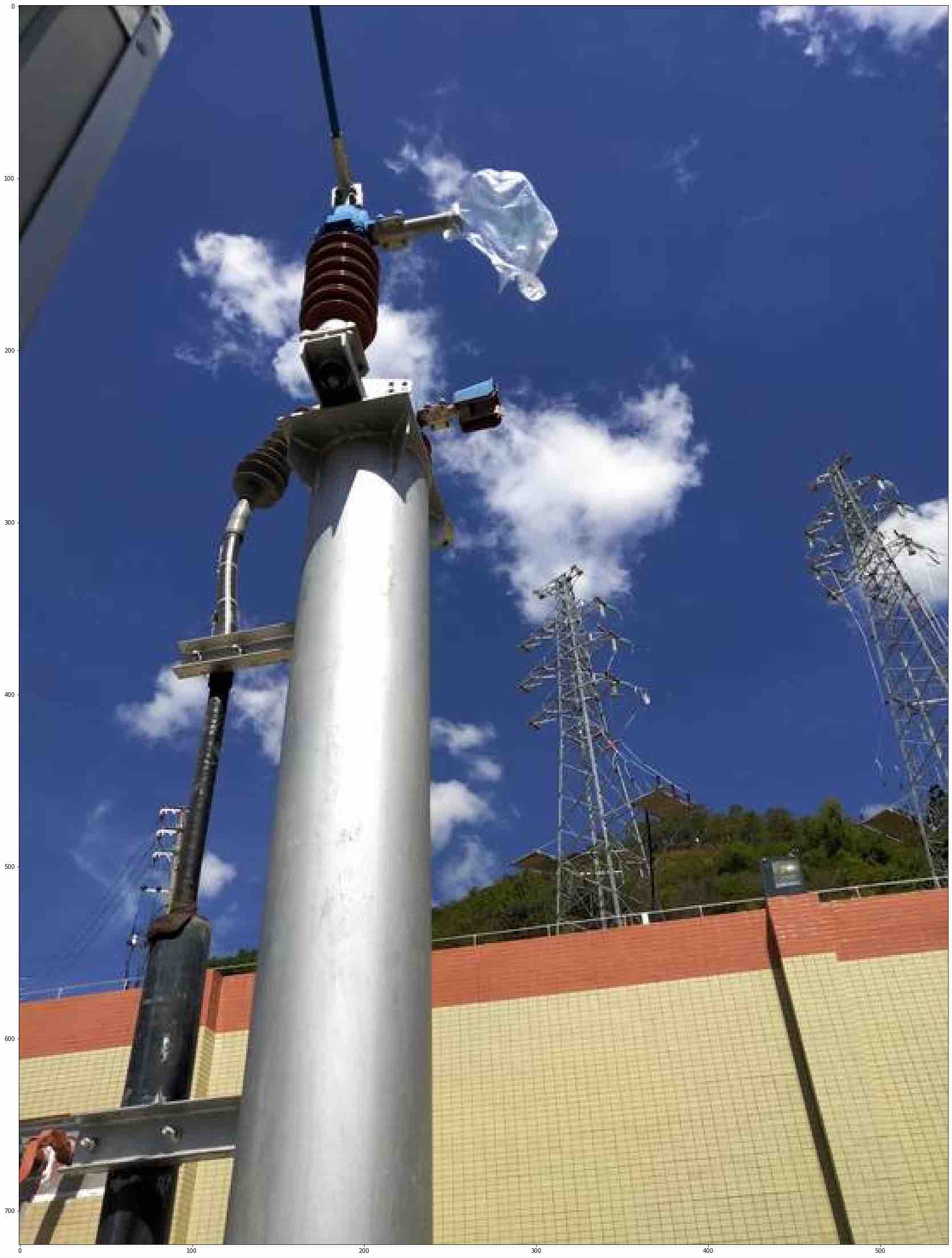}
      \end{subfigure}
      \hfill
      \begin{subfigure}[t]{0.49\linewidth}
        \includegraphics[width=\linewidth]{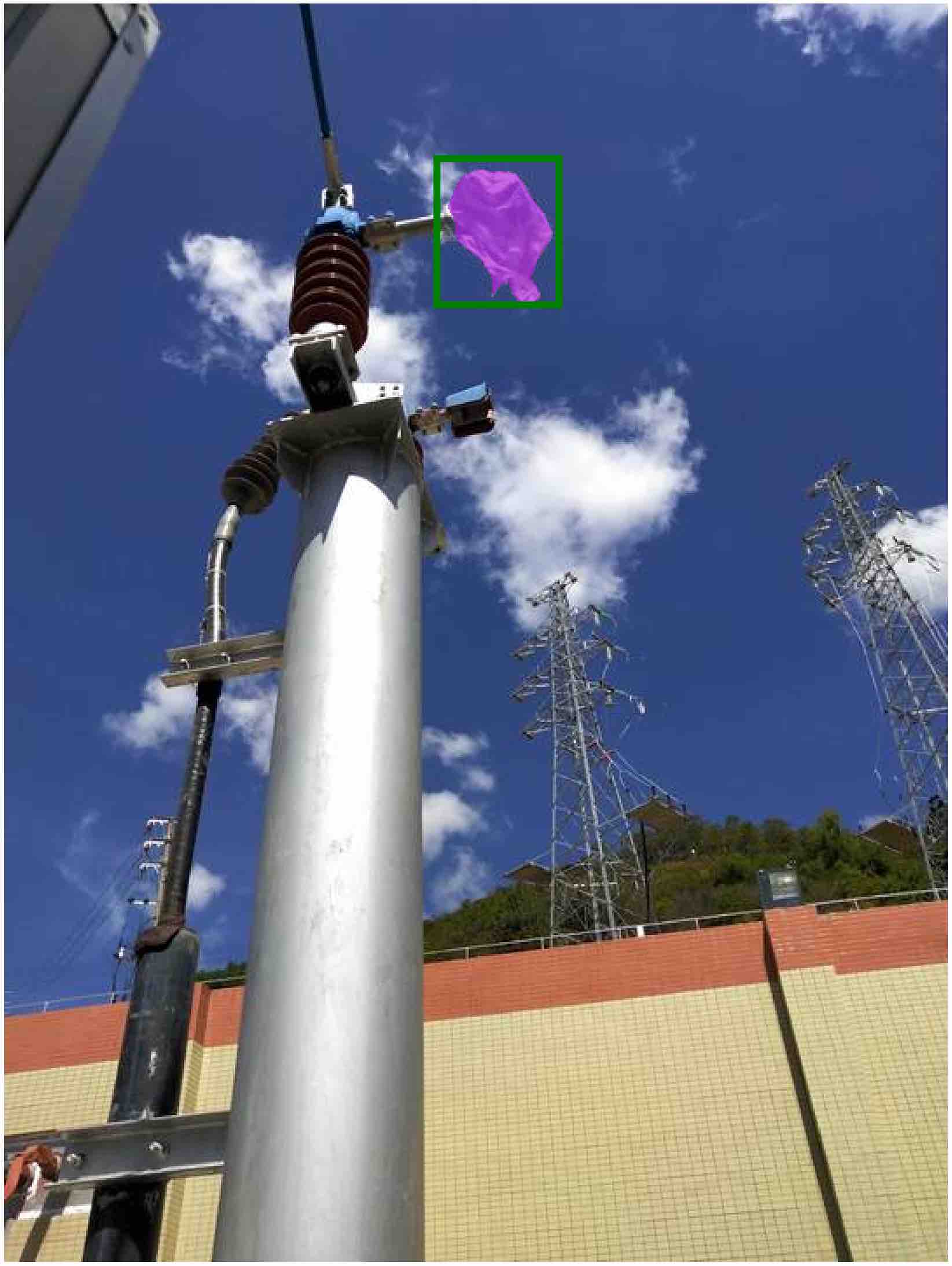}
      \end{subfigure}
      \caption{Identify any anomaly object}
      \label{fig:cases_7}
  \end{subfigure}
  \\
  \begin{subfigure}[t]{0.65\linewidth}
      \begin{subfigure}[t]{0.49\linewidth}
        \includegraphics[width=\linewidth]{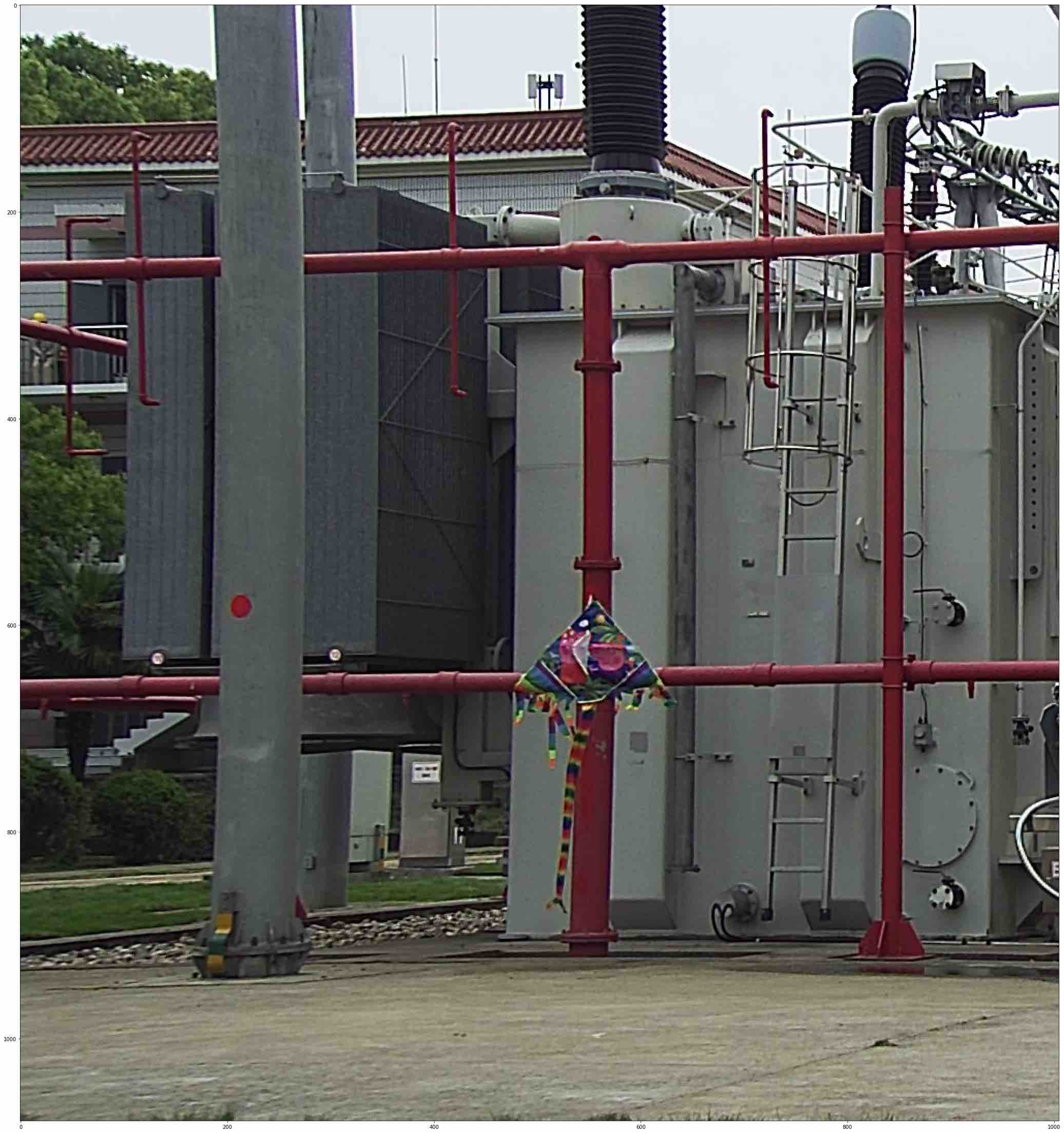}
      \end{subfigure}
      \hfill
      \begin{subfigure}[t]{0.49\linewidth}
        \includegraphics[width=\linewidth]{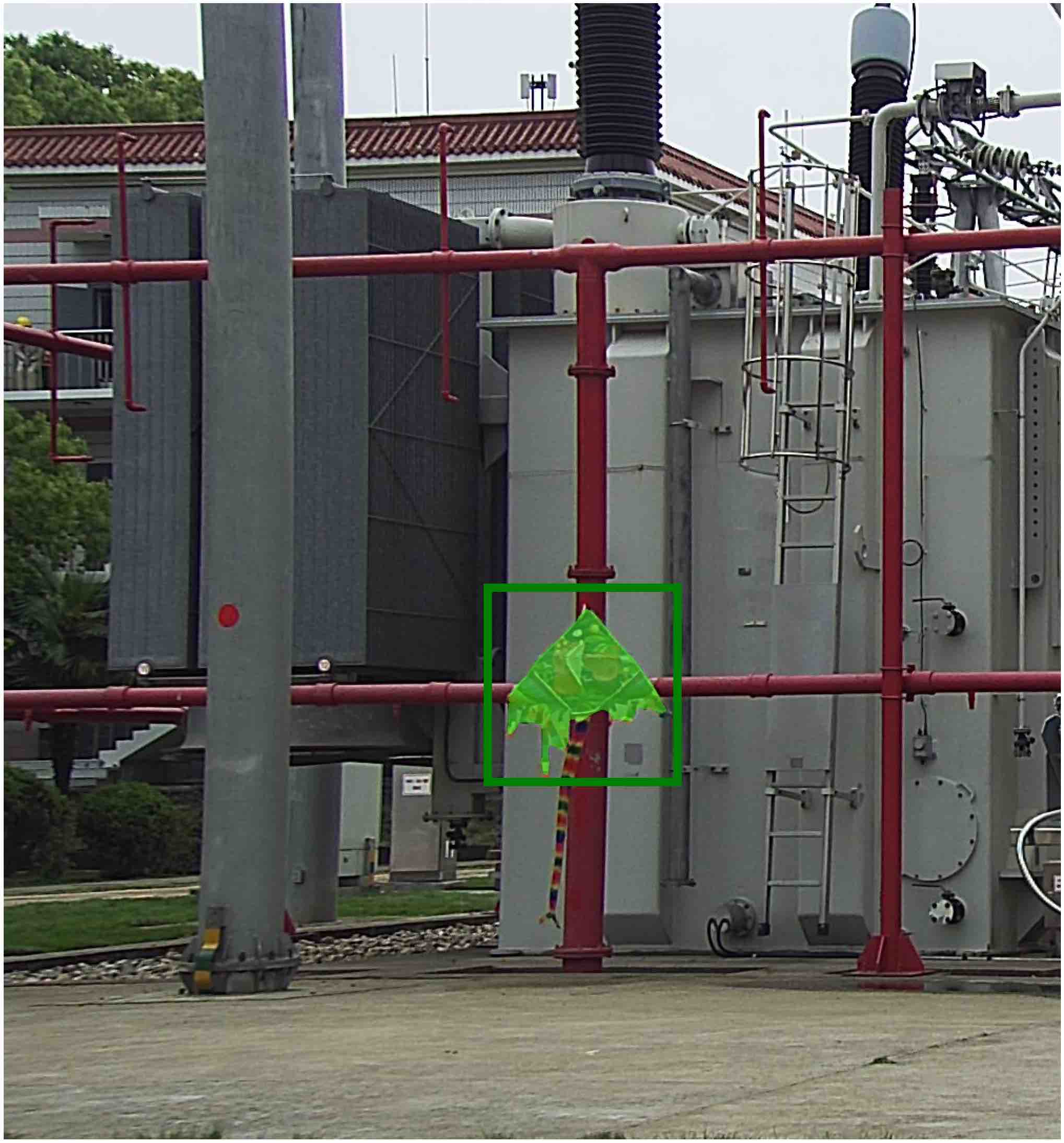}
      \end{subfigure}
      \caption{Find any anomaly object and highlight it}
      \label{fig:cases_8}
  \end{subfigure}
  \caption{Text-conditioned image understanding results. For each subfigure, the caption and the left image are user inputs whereas the right image is the final response of VisionGPT. As we can see, VisionGPT can follow user requirements to detect and segment correct objects, even though in some cases, i.e., (f), (g), and (h), users do not specify an object name.}
  \label{fig:cases}
\end{figure}

\section{Experiments}
\label{sec:experiments}
In this section, we showcase how VisionGPT address different tasks by integrating state-of-the-art foundation models.

\paragraph{Text-Conditioned Image Understanding.} By integrating the intelligence and capacities of Llama 2\cite{llama2}, YOLO series\cite{supergradients,Jocher_Ultralytics_YOLO_2023}, and SAM\cite{kirillov2023segment}, our VisionGPT can highlight the desired object in complex realistic scenes (\cref{fig:cases_1,fig:cases_2,fig:cases_3}) and animation scenes (\cref{fig:cases_4}), achieve instance segmentation (\cref{fig:cases_5}), and detect correct objects even without given a specific object name (\cref{fig:cases_6,fig:cases_7,fig:cases_8}).

\paragraph{Text-Conditioned Image Generation/Editing.} As shown in \cref{fig:g_cases}, our VisionGPT can edit the given images based on user requests through the additional integration of Stable Diffusion models\cite{rombach2022high,stable}. 

\begin{figure}[htbp]
  \centering
  \begin{subfigure}[t]{0.8\linewidth}
      \begin{subfigure}[t]{0.49\linewidth}
        \includegraphics[width=\linewidth]{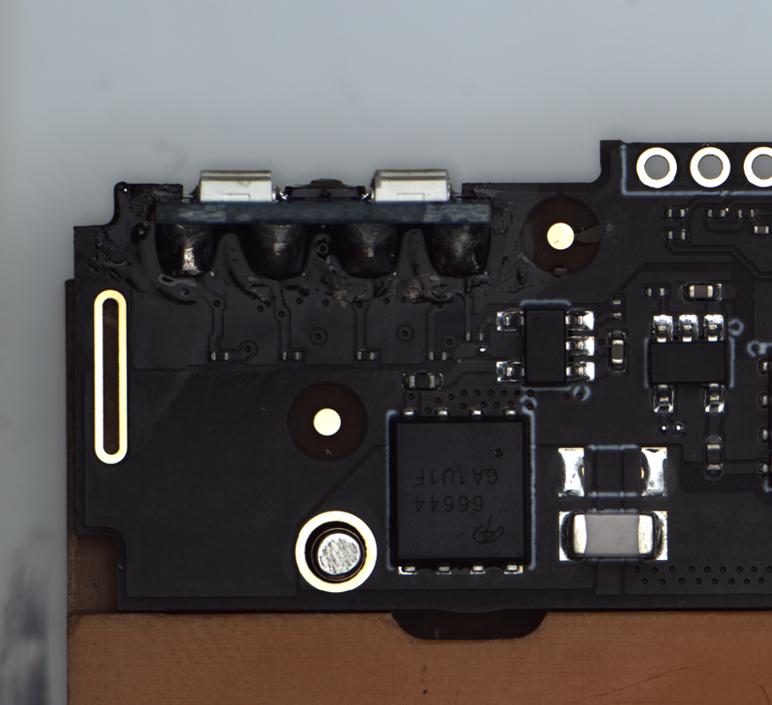}
      \end{subfigure}
      \hfill
      \begin{subfigure}[t]{0.49\linewidth}
        \includegraphics[width=\linewidth]{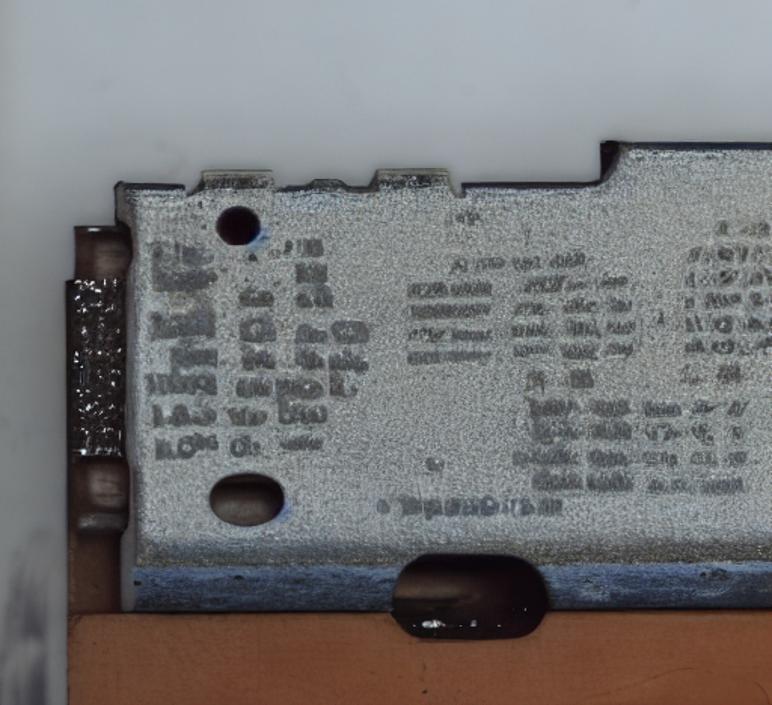}
      \end{subfigure}
      \caption{Please cover the naked capacitors}
      \label{fig:g_cases_1}
  \end{subfigure}
  \\
  \begin{subfigure}[t]{0.8\linewidth}
      \begin{subfigure}[t]{0.49\linewidth}
        \includegraphics[width=\linewidth]{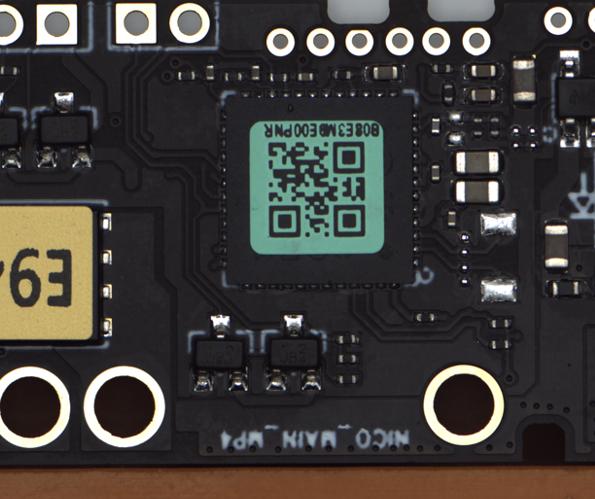}
      \end{subfigure}
      \hfill
      \begin{subfigure}[t]{0.49\linewidth}
        \includegraphics[width=\linewidth]{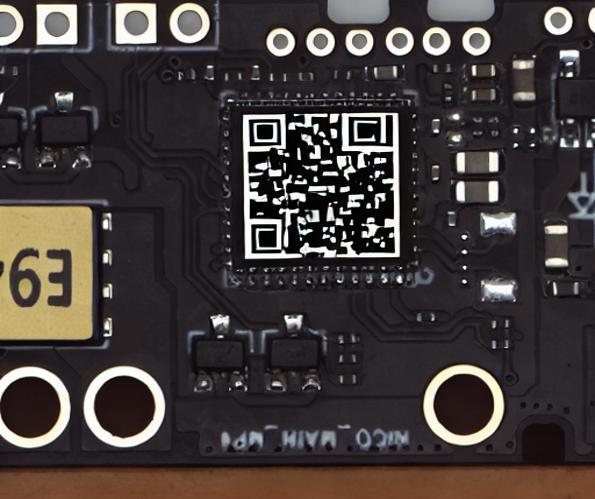}
      \end{subfigure}
      \caption{Find the QR Code and generate a new one}
      \label{fig:g_cases_2}
  \end{subfigure}
  \\
  \begin{subfigure}[t]{0.8\linewidth}
      \begin{subfigure}[t]{0.49\linewidth}
        \includegraphics[width=\linewidth]{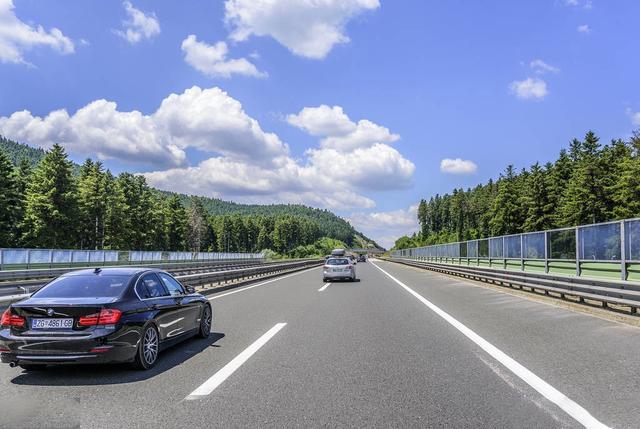}
      \end{subfigure}
      \hfill
      \begin{subfigure}[t]{0.49\linewidth}
        \includegraphics[width=\linewidth]{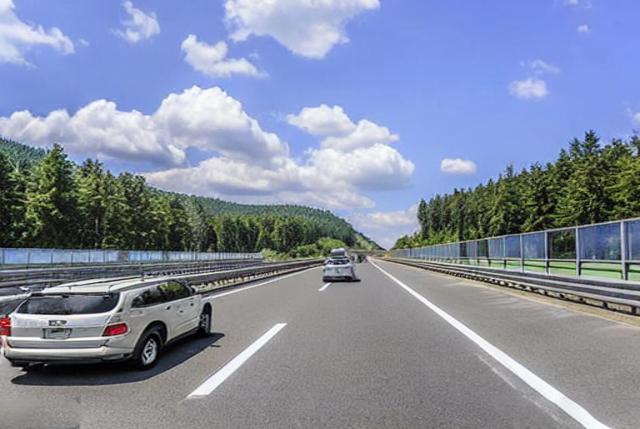}
      \end{subfigure}
      \caption{Replace the front car with a different type}
      \label{fig:g_cases_3}
  \end{subfigure}
  \\
  \begin{subfigure}[t]{0.8\linewidth}
      \begin{subfigure}[t]{0.49\linewidth}
        \includegraphics[width=\linewidth]{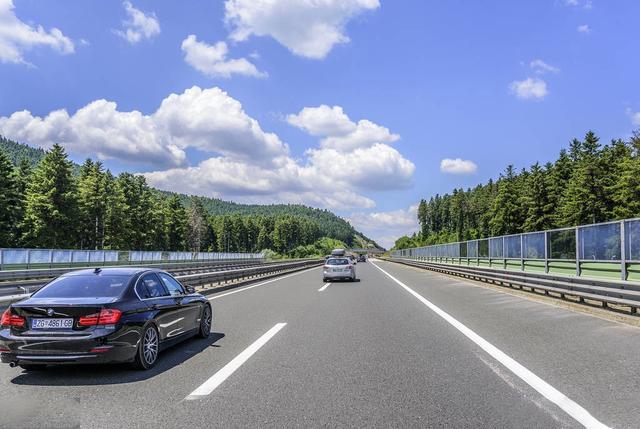}
      \end{subfigure}
      \hfill
      \begin{subfigure}[t]{0.49\linewidth}
        \includegraphics[width=\linewidth]{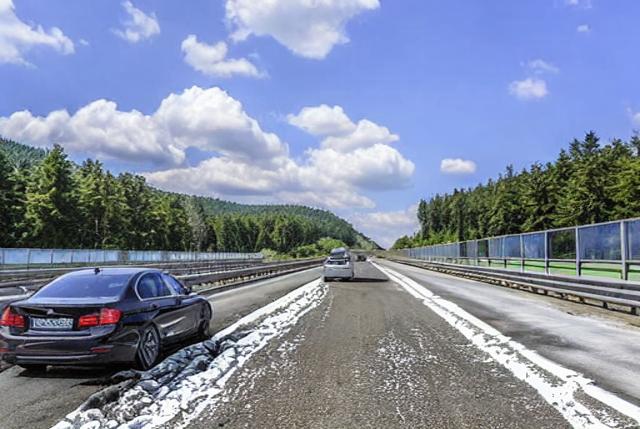}
      \end{subfigure}
      \caption{Please have a icy highway}
      \label{fig:g_cases_4}
  \end{subfigure}
\end{figure}

\begin{figure}[htbp]
  \ContinuedFloat
  \centering
  \begin{subfigure}[t]{0.8\linewidth}
      \begin{subfigure}[t]{0.49\linewidth}
        \includegraphics[width=\linewidth]{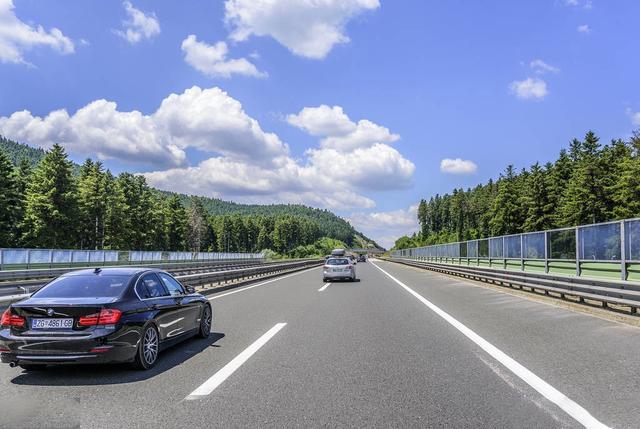}
      \end{subfigure}
      \hfill
      \begin{subfigure}[t]{0.49\linewidth}
        \includegraphics[width=\linewidth]{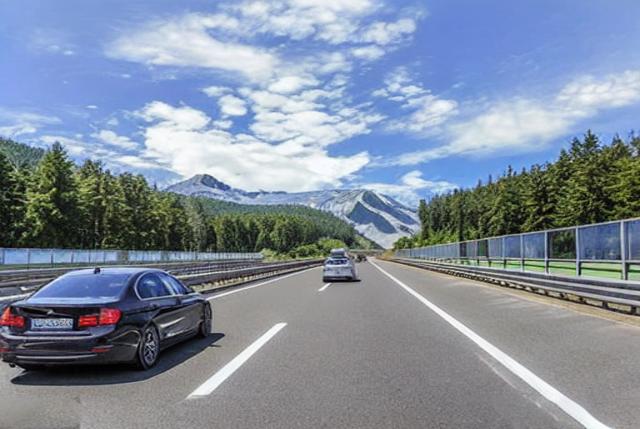}
      \end{subfigure}
      \caption{Please generate a mountain far away}
      \label{fig:g_cases_5}
  \end{subfigure}
  \\
  \begin{subfigure}[t]{0.8\linewidth}
      \begin{subfigure}[t]{0.49\linewidth}
        \includegraphics[width=\linewidth]{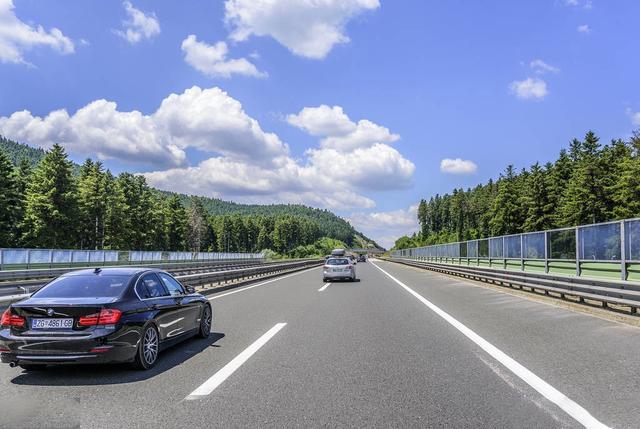}
      \end{subfigure}
      \hfill
      \begin{subfigure}[t]{0.49\linewidth}
        \includegraphics[width=\linewidth]{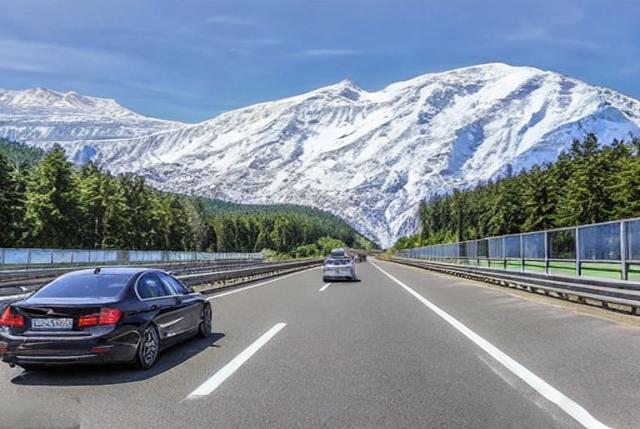}
      \end{subfigure}
      \caption{A big mountain at the end of the road}
      \label{fig:g_cases_6}
  \end{subfigure}
  \\
  \begin{subfigure}[t]{0.8\linewidth}
      \begin{subfigure}[t]{0.49\linewidth}
        \includegraphics[width=\linewidth]{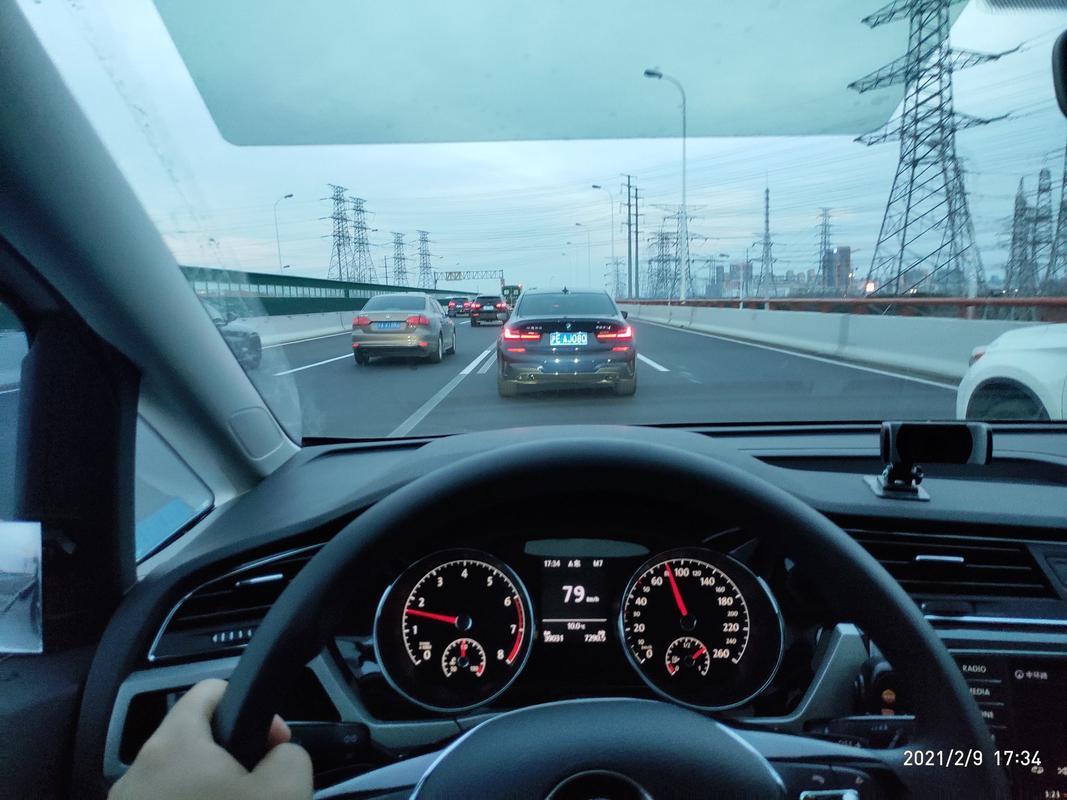}
      \end{subfigure}
      \hfill
      \begin{subfigure}[t]{0.49\linewidth}
        \includegraphics[width=\linewidth]{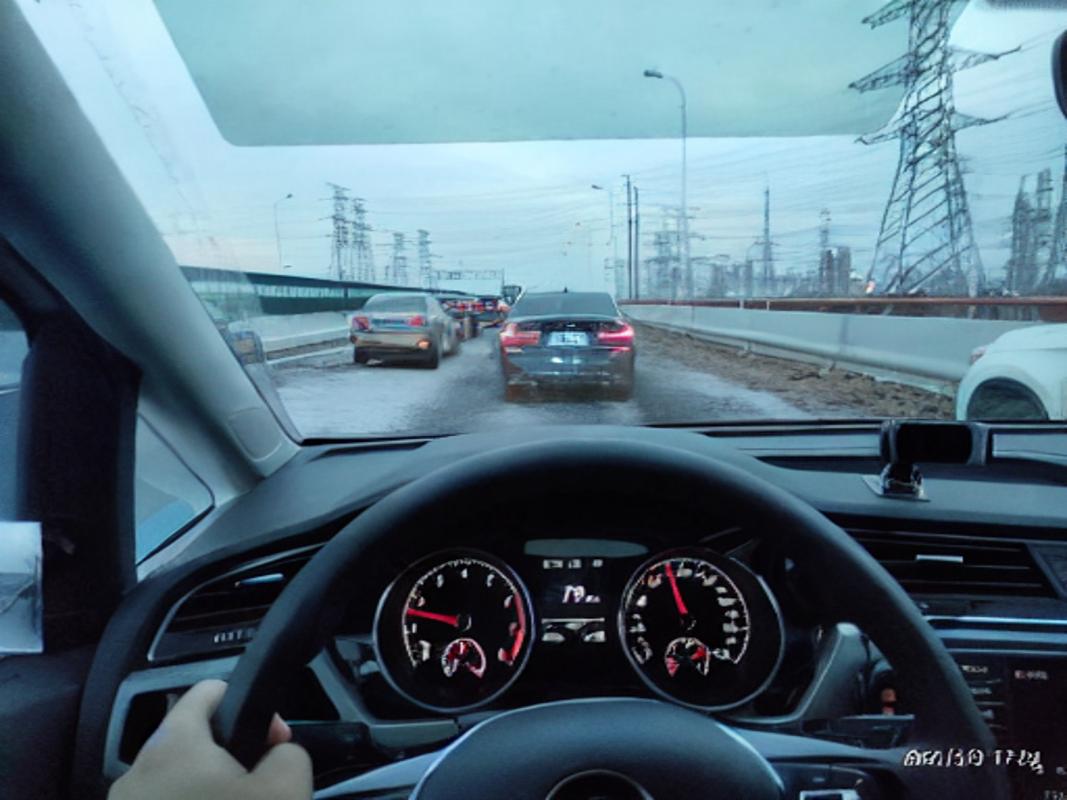}
      \end{subfigure}
      \caption{It is so cold outside}
      \label{fig:g_cases_7}
  \end{subfigure}
  \\
  \begin{subfigure}[t]{0.8\linewidth}
      \begin{subfigure}[t]{0.49\linewidth}
        \includegraphics[width=\linewidth]{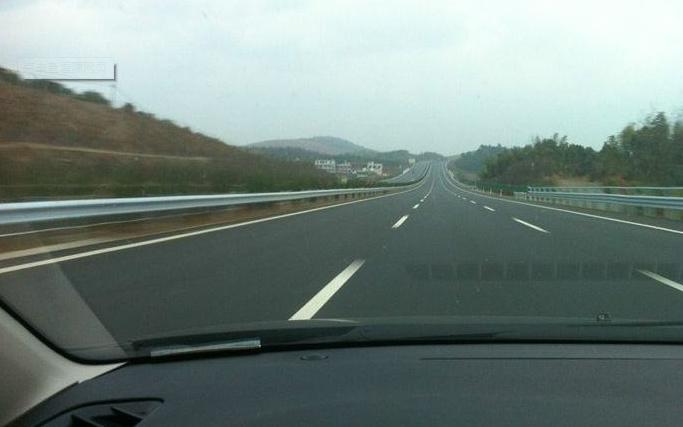}
      \end{subfigure}
      \hfill
      \begin{subfigure}[t]{0.49\linewidth}
        \includegraphics[width=\linewidth]{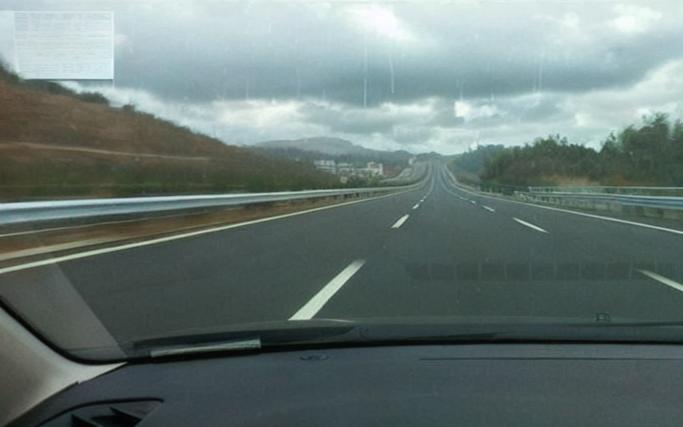}
      \end{subfigure}
      \caption{It is going to rain soon}
      \label{fig:g_cases_8}
  \end{subfigure}
  \caption{Text-conditioned image generation/editing results. For each subfigure, the caption and the left image are user inputs whereas the right image is the final response of VisionGPT. As we can see, VisionGPT can generate image outputs aligned with user custom requests.}
  \label{fig:g_cases}
\end{figure}

\section{Limitations}

While VisionGPT presents a substantial advancement in integrating state-of-the-art (SOTA) computer vision models with large language models (LLMs), it is not without its limitations. A primary constraint lies in the rapid evolution of SOTA vision foundation models and vision-language models, which poses a challenge to maintaining the relevance and effectiveness of VisionGPT. The continual emergence of new models and techniques necessitates frequent updates and adaptations to the VisionGPT framework to ensure compatibility and optimal performance. Another limitation stems from VisionGPT's reliance on the integration of multiple expert models. While this design enhances versatility and ensures the utilization of the best features from various models, it also introduces complexity in model management and coordination. To ensure seamless interaction and data flow between frequently evolving models, meticulous design and maintenance are required. Additionally, while VisionGPT aims to act as a vision-language model, its performance is contingent on the integrated vision models' and LLMs' quality and capabilities. Any shortcomings or biases in these underlying models could potentially propagate through the VisionGPT framework, affecting the overall performance and reliability of the system.

In conclusion, although VisionGPT represents a significant stride toward a unified and versatile AI system, it is not exempt from limitations. Addressing the seamless integration and management of diverse expert models, alongside navigating the challenges presented by rapid advancements in vision foundation models, underscores crucial areas for future enhancement and development.

\section{Conclusion}

In this work, we have presented VisionGPT, a novel agent that integrates state-of-the-art large language models (LLMs) with vision foundation models. The integration of LLMs enables VisionGPT to extract context, details, and intent from user inputs, translating them into precise action proposals, whereas vision foundation models can collaborate within VisionGPT to achieve better efficiency, versatility, and performance. Therefore, VisionGPT automates the entire workflow from request understanding to response generation, offering a robust and adaptable platform for vision-language understanding and various vision-oriented AI tasks. Looking ahead, there lies vast potential for VisionGPT to evolve and seamlessly integrate with forthcoming LLMs and vision models, offering even more personalized and context-aware interactions. This research establishes the foundation for future advancements in this field, to consistently enhance and customize AI systems to address the diverse and expanding requirements of users.


%
%
\bibliographystyle{splncs04}
\bibliography{refs}
\end{document}